\documentclass{article}

\PassOptionsToPackage{numbers, compress}{natbib}
\usepackage[preprint]{neurips_2024}
\usepackage[utf8]{inputenc} % allow utf-8 input
\usepackage[T1]{fontenc}    % use 8-bit T1 fonts
\usepackage{hyperref}       % hyperlinks
\usepackage{url}            % simple URL typesetting
\usepackage{booktabs}       % professional-quality tables
\usepackage{amsfonts}       % blackboard math symbols
\usepackage{nicefrac}       % compact symbols for 1/2, etc.
\usepackage{microtype}      % microtypography
\usepackage{xcolor}         % colors

\usepackage{amssymb}
\usepackage{graphicx}
\usepackage{amsmath}
\usepackage{subcaption}

\usepackage{wrapfig}

\newtheorem{theorem}{Theorem}

\newtheorem{definition}{Definition}

\newtheorem{proposition}{Proposition}

\newcommand{\myendofproof}[0]{\hfill $\blacksquare$ \newline}

\title{New Desiderata for Direct Preference Optimization}

\author{%
  Xiangkun Hu \hspace*{0.4cm} Tong He \hspace*{0.4cm} David Wipf\\
  Amazon Web Services \\
  \texttt{\{xiangkhu,htong,daviwipf\}@amazon.com} \\
}

\begin{document}

\maketitle

\input{def.set}

\vspace*{-0.4cm}
\begin{abstract}
Large language models in the past have typically relied on some form of reinforcement learning with human feedback (RLHF) to better align model responses with human preferences.  However, because of oft-observed instabilities when implementing these RLHF pipelines, various reparameterization techniques have recently been introduced to sidestep the need for separately learning an RL reward model.  Instead, directly fine-tuning for human preferences is achieved via the minimization of a single closed-form training objective, a process originally referred to as direct preference optimization (DPO) and followed by several notable descendants.  Although effective in certain real-world settings, we introduce new evaluation criteria that serve to highlight unresolved shortcomings in the ability of existing DPO methods to interpolate between a pre-trained reference model and empirical measures of human preferences, as well as unavoidable trade-offs in how low- and high-quality responses are regularized and constraints are handled.  Our insights then motivate an alternative DPO-like loss that provably mitigates these limitations.  Empirical results  serve to corroborate notable aspects of our analyses\footnote{A preliminary version of this work appears in the ICML 2024 Workshop on Models of Human Feedback for AI Alignment: \url{https://openreview.net/pdf?id=Fgf0iAOb22}}.
\end{abstract}

\vspace*{-0.3cm}
\section{Introduction}
\vspace*{-0.2cm}
Although pre-trained large language models (LLMs) often display remarkable capabilities \cite{bubeck2023sparks, chang2024survey,  openai2024gpt4, LLMSurvey}, it is well-established that they are prone to responding in ways that may be at odds with human preferences for rationale discourse \cite{bai2022constitutional, gallegos2023bias}.  To this end, after an initial supervised fine-tuning phase that produces a reference model or policy $\pi_{\tiny \mbox{ref}}(y|x)$, it is now commonplace to apply reinforcement learning with human feedback (RLHF) to further refine the LLM responses $y$ to input prompts $x$ \cite{ziegler2019fine, stiennon2009learning, bai2022training, ouyang2022training}. This multi-step process involves first learning a reward model that reflects human inclinations culled from labeled preference data, and then subsequently training a new policy that balances reward maximization with proximity to $\pi_{\tiny \mbox{ref}}(y|x)$.

Because RLHF introduces additional complexity, computational overhead, and entry points for instability, clever reparameterization techniques have recently been proposed that sidestep the need for separately learning a reward model altogether.  Instead, increased alignment with human preferences is achieved via the minimization of a single closed-form training objective, a process originally referred to as direct preference optimization (DPO) \cite{rafailov2024direct} followed by several notable descendants and generalizations \cite{azar2024general,tang2024generalized,wang2024beyond,zhao2023slic}.  These alternatives dramatically economize model development; however, with recency comes the potential that the consequences of less obvious properties of DPO-based objectives may still be under-explored.  It is along these lines that our attention herein lies, with the end goal of quantifying and steering model behavior in transparently beneficial directions.

% with the end goal of quantifying and later improving behavior through the lens of targeted desiderata.

% for blindspots in their application and the possibility of 

% because new, not fully understood.

% These alternatives are attractive

% alternative approaches to human preference optimization have recently been proposed that sidestep the need for learning a reward model altogether.

% First, a reward model must be trained separately using labeled preference data; subsequently, a new parameterized policy $\pi_\theta(y|x)$ is optimized via RL to balance maximization of the reward $\pi_{\tiny \mbox{ref}}(y|x)$

After introducing basic concepts and the details of existing preference optimization models in Section \ref{sec:backgroun}, the remainder of the paper devoted to our technical contributions can be distilled as follows:
\begin{itemize}
    \item We introduce new evaluation desiderata that comport with intuition regarding how a preference model ideally should behave, and yet (somewhat surprisingly) are provably \textit{not} satisfied by a broad class of existing DPO-based approaches.  In particular, we show that because of uniform regularization effects, the minimizers of commonly-used preference optimization objectives like DPO are at times unable to  \textit{preserve} performance in regions where the reference model is strong while \textit{simultaneously} improving upon the reference model elsewhere (Section \ref{sec:equiv_criteria_analysis}).  Moreover, we also elucidate limitations in the ability to \textit{interpolate} between ideal endpoints as model trade-off parameters are varied (Section \ref{sec:interpolate_criteria_analysis}).
    
\item We prove that once inevitable learning constraints are introduced (explicitly or implicitly, e.g., early-stopping, weight decay, etc.), the core reparameterizations that underpin certain DPO models no longer strictly hold (Section \ref{sec:constraints}).  This motivates alternative justifications based solely on properties of the final loss functions involved (Appendices \ref{sec:dpo_noise_adaptive_loss} and \ref{sec:gaussian_dpo}).

% % there exist unavoidable trade-offs
%     \item constraints
    
\item Based on the above, we introduce a new preference optimization loss called $\ell_{\tiny \mbox{TYPO}}$ that, by design, satisfies our evaluation desiderata while avoiding any dependency on constraint-dependent reparameterizations (Section \ref{sec:new_objectives}).  Properties of this loss relative to its precursors are also corroborated using Monte-Carlo simulations  (Section \ref{sec:experiments} and Appendix \ref{app:additional_interpolation_results}).

\end{itemize}

\vspace*{-0.2cm}
\section{Background} \label{sec:backgroun}
\vspace*{-0.2cm}
% As context for the analysis that follows in later sections, we first present preliminaries of human preference optimization followed by the basics of RLHF, DPO, and IPO\tong{cite for the first appearance?} \david{Right, should do so if this part stays.  But I think for space will just cut this whole introductory sentence.} formulations and extensions thereof.

% \subsection{Preliminaries}

We adopt $x \sim \calD_x$ to denote an \textit{input prompt} $x$ drawn from some distribution $\calD_x$.  From here, conditioned on such prompts we may then generate \textit{responses} $y$ using a pre-trained reference language model/policy $\pi_{\tiny \mbox{ref}}(y|x)$.  Moreover, given a pair of such responses $y_1 \neq y_2$,  we adopt the binary indicator variable $z = \mathbb{I}[y_1 \succ y_2| y_1, y_2, x ]$ to convey that $y_1$ is preferred over $y_2$ by a human evaluator when $z = 1$, or else $z = 0$ if instead $y_2 \succ y_1$.  Given a population of such evaluators, we express the ground-truth human preference distribution as $p^*(z|y_1,y_2,x) = p^*(y_1 \succ y_2|y_1,y_2,x)$.  And finally, we define a set of human labeled tuples  drawn from a training distribution $\calD_{tr}$ as
\begin{equation} \label{eq:preference_sampling}
    \{y_w,y_l,x\} \sim \calD_{tr} ~~ \equiv ~~ \{z,y_1,y_2,x\} \sim \calD_{tr} ~~\equiv ~~ z \sim p^*(z|y_1,y_2,x), ~\{y_1,y_2\} \sim \pi_{\tiny \mbox{ref}}(y|x), x \sim \calD_x,
\end{equation}
where $y_w \succ y_l$ (subscripts here stand for `win' and `lose').\footnote{We generally assume that $y_1 \neq y_2$; however, the $y_1 = y_2$ case can nonetheless be handled by simply assigning $p^*(z|y,y,x) = 1/2$, inclusion of which does not effect the analysis that follows.  In particular, such cases merely introduce an irrelevant constant into the human preference loss functions under consideration.}  In other words, each training tuple is generated by drawing $x$ from $\calD_x$, $y_1 \neq y_2$ from the reference policy $\pi_{\tiny \mbox{ref}}$, and finally $z$ is produced by human labelers that operate according to $p^*$.  Note that per convention in prior work and ease of presentation, we will often abbreviate the preference distribution notation as $p^*(y_1 \succ y_2|y_1,y_2,x) \equiv p^*(y_1 \succ y_2|x )$ when the context is sufficiently clear.

\subsection{Reinforcement Learning with Human Feedback (RLHF)}

\paragraph{Reward Function Estimation:}~~Given two candidate responses $y_1 \neq y_2$  sampled using prompt $x$, the Bradley-Terry (BT) model \cite{bradley1952rank} for human preferences stipulates that
\begin{equation} \label{eq:BT}
    p^*(y_1 \succ y_2|x) = \frac{\exp[r^*(y_1,x)]}{\exp[r^*(y_1,x)] + \exp[r^*(y_2,x)]} = \sigma\big[ r^*(y_1,x) - r^*(y_2,x)   \big],
\end{equation}
where $r^*(y,x)$ is a so-called latent reward model and $\sigma$ is the logistic function.  Because $r^*(y,x)$ is unobservable, it is not possible to directly compute $p^*(y_1 \succ y_2|x)$; however, we can train an approximation  $p_\phi(y_1 \succ y_2|x)$ (equivalent to $p_\phi(y_1 \succ y_2|y_1,y_2,x)$ as before) defined by a parameterized proxy reward $r_\phi(y,x)$.  Specifically, we can minimize the loss
\begin{equation}
\ell_{\tiny \mbox{BT}}(r_\phi) :=  \mathbb{E}_{\{y_w,y_l,x\} \sim \calD_{\tiny \mbox{tr}}} \Big[ -\log p_\phi(y_w \succ y_l|x) \Big] = \mathbb{E}_{\{y_w,y_l,x\} \sim \calD_{\tiny \mbox{tr}}} \Big[ -\log \sigma\big[ r_\phi(y_w,x) - r_\phi(y_l,x)   \big] \Big].
\end{equation}
The optimized reward $\hat{r}_\phi(y,x) := \arg\min_{r_\phi} \ell_{\tiny \mbox{BT}}(r_\phi) \approx r^*(y,x)$ can then be applied to fine-tuning the pre-trained reference model $\pi_{\tiny \mbox{ref}}(y|x)$ as described next.

% \tong{$\pi_\phi$ is not defined anywhere, I think it worth to clarify its independence to $\pi_\theta$} \david{Good catch.  Actually, this is a typo, it should be $r_\phi$ (not $\pi_\phi$) which has been defined.}

%\vspace*{0.3cm}
\paragraph{RL Fine-Tuning with Estimated Reward Function:}~~The goal here is to improve upon a given $\pi_{\tiny \mbox{ref}}(y|x)$ using a separate trainable model $\pi_{\theta}(y|x)$, the high-level desiderata being: (i) Maximize the previously-estimated reward function $\hat{r}_{\phi}(y,x)$ when following $\pi_{\theta}(y|x)$, while (ii) Minimizing some measure of distance between $\pi_{\theta}(y|x)$ and $\pi_{\tiny \mbox{ref}}(y|x)$ to avoid overfitting merely to preference rewards.  These objectives typically materialize through the minimization of
\begin{equation} \label{eq:rlhf_loss}
\ell_{\tiny \mbox{RLHF}}\left(\pi_\theta, \pi_{\tiny \mbox{ref}}, \hat{r}_\phi, \lambda \right) ~~ := ~~ \mathbb{E}_{y \sim \pi_{\theta}(y|x), x\sim \calD_x} \Big[ -\hat{r}_\phi(y,x) \Big] + \lambda ~\mathbb{E}_{x\sim \calD_x} \Big[ \mathbb{KL}\big[\pi_\theta(y|x)  || \pi_{\tiny \mbox{ref}}(y|x)  \big] \Big],
\end{equation}
where $\lambda > 0$ is a trade-off parameter.  Although not differentiable, starting from an initialization such as $\pi_\theta = \pi_{\tiny \mbox{ref}}$, the loss $\ell_{\tiny \mbox{RLHF}}\left(\pi_\theta, \pi_{\tiny \mbox{ref}}, \hat{r}_\phi, \lambda \right)$ can be optimized over $\pi_\theta$ using various forms of RL~\cite{schulman2017proximal, ramamurthy2022reinforcement}

\subsection{Direct Preference Optimization (DPO)}
\vspace*{-0.2cm}

Consider now the reward-dependent RLHF loss $\ell_{\tiny \mbox{RLHF}}$ from (\ref{eq:rlhf_loss}) defined w.r.t.~and arbitrary reward function $r(y,x)$.  DPO \cite{rafailov2024direct} is based on the observation that, provided $\pi_\theta$ is sufficiently flexible such that we may treat it as an arbitrary function for optimization purposes,\footnote{This is a key assumption with non-trivial consequences; Section \ref{sec:constraints} will explore this issue in further detail.} the minimum of $\ell_{\tiny \mbox{RLHF}}\left(\pi_\theta, \pi_{\tiny \mbox{ref}}, r, \lambda \right)$ w.r.t.~$\pi_\theta$ can be directly computed as
\begin{equation} \label{eq:closed_form_rlhf_policy}
    \pi_r(y|x) ~:=~ \arg\min_{\pi_\theta} \ell_{\tiny \mbox{RLHF}}\left(\pi_\theta, \pi_{\tiny \mbox{ref}}, r, \lambda \right) = \frac{1}{Z(x)}\pi_{\tiny \mbox{ref}}(y|x) \exp\left[\frac{1}{\lambda} r(y,x)  \right],
\end{equation}
where $Z(x) := \sum_y \pi_{\tiny \mbox{ref}}(y|x) \exp\left[\frac{1}{\lambda} r(y,x)  \right]$ is the partition function ensuring that $\pi_r(y|x)$ forms a proper distribution \cite{peng2019advantage,peters2007reinforcement}.  From here, assuming $\pi_{\tiny \mbox{ref}}(y|x) > 0$, we can rearrange (\ref{eq:closed_form_rlhf_policy}) to equivalently establish that 
\begin{equation} \label{eq:closed_form_rlhf_reward}
    r(y,x) = \lambda \log \frac{ \pi_r(y|x) }{\pi_{\tiny \mbox{ref}}(y|x)} + \lambda \log Z(x).
\end{equation}
Because thus far $r$ has remained unspecified, it naturally follows that these policy/reward relationships hold even for the ground-truth reward $r^*$ and the associated optimal policy $\pi^{**}(y|x) := \arg\min_{\pi_\theta} \ell_{\tiny \mbox{RLHF}}\left(\pi_\theta, \pi_{\tiny \mbox{ref}}, r^*, \lambda \right)$. Hence instead of approximating $r^*(y,x)$ with $r_\phi(y,x)$ as in (\ref{eq:BT}), we may equivalently approximate $\pi^{**}(y|x)$ with some $\pi_\theta(y|x)$ leading to the  DPO loss ~~ $\ell_{\tiny \mbox{DPO}}(\pi_\theta, \pi_{\tiny \mbox{ref}},\lambda) ~:= $
\begin{equation} \label{eq:dpo_loss}
\ell_{\tiny \mbox{BT}}\left( \lambda \log \frac{ \pi_\theta(y|x) }{\pi_{\tiny \mbox{ref}}(y|x)} \right) = \mathbb{E}_{\{y_w,y_l,x\} \sim \calD_{\tiny \mbox{tr}}} \left[ -\log \sigma\left( \lambda \log \frac{ \pi_\theta(y_w|x) }{\pi_{\tiny \mbox{ref}}(y_w|x)} - \lambda \log \frac{ \pi_\theta(y_l|x) }{\pi_{\tiny \mbox{ref}}(y_l|x)}   \right) \right],
\end{equation}
noting that the partition function $Z(x)$ conveniently cancels out and can be excluded from further consideration.  It is now possible to directly optimize (\ref{eq:dpo_loss}) over $\pi_\theta$ using SGD without the need for any challenging RLHF procedure.  The basic intuition here is that the parameterized policy $\pi_\theta$ induces an implicit reward $\lambda \log \left[ \pi_\theta(y|x)  \pi_{\tiny \mbox{ref}}^{-1}(y|x) \right]$ that is being optimized via the original BT preference model.  Moreover this equivalence is exact assuming data distributed as in (\ref{eq:preference_sampling}).

\vspace*{-0.1cm}
\subsection{Identity Preference Optimization (IPO)}
\vspace*{-0.1cm}
Similar to DPO, the identity preference optimization (IPO) formulation  \cite{azar2024general} avoids both a 2-step learning process and cumbersome, potentially unstable RL training.  To accomplish this, IPO is predicated on minimizing the original RLHF loss from (\ref{eq:rlhf_loss}) but with an alternative reward function.  Specifically, the motivating IPO objective is to minimize $\ell_{\tiny \mbox{RLHF}}\left(\pi_\theta,\pi_{\tiny \mbox{ref}}, r_{\tiny \mbox{IPO}}, \lambda \right)$, where
\begin{equation} \label{eq:ipo_reward}
r_{\tiny \mbox{IPO}}(y,x) := \mathbb{E}_{y' \sim \pi_{\tiny \mbox{ref}}(y|x)} \big[ p^*(y \succ y' |x,y,y') \big],
\end{equation}
over $\pi_\theta$.\footnote{Note that in principle the distribution used to draw samples $y'$ in defining $r_{\tiny \mbox{IPO}}$ need not be set to $\pi_{\tiny \mbox{ref}}$; however, in practice $\pi_{\tiny \mbox{ref}}$ is a typical choice, which we adopt throughout for simplicity.}  Because of the special structure of \textit{this particular} reward function, it turns out that it is possible to minimize $\ell_{\tiny \mbox{RLHF}}\left(\pi_\theta,\pi_{\tiny \mbox{ref}}, r_{\tiny \mbox{IPO}}, \lambda \right)$ over $\pi_\theta$ without RL.  In brief, this is accomplished by first noting that for any pair of responses $y_1 \neq y_2$ the existence of an optimal IPO policy, denoted $\pi_{\tiny \mbox{IPO}}$, evaluated at these responses can be computed as a function of the reward $r_{\tiny \mbox{IPO}}$ using (\ref{eq:closed_form_rlhf_policy}).  Combining $y_1$ and $y_2$ dependent terms, after a few algebraic manipulations this then leads to the equivalence relation
\begin{equation} \label{eq:ipo_equivalence_relation}
    \log \left[ \frac{\pi_{\tiny \mbox{IPO}}(y_1|x) \pi_{\tiny \mbox{ref}}(y_2|x)}{\pi_{\tiny \mbox{IPO}}(y_2|x) \pi_{\tiny \mbox{ref}}(y_1|x)}\right] = \frac{1}{\lambda}\big[ r_{\tiny \mbox{IPO}}(y_1,x) - r_{\tiny \mbox{IPO}}(y_2,x) \big].
\end{equation}
However, unlike DPO where an analogous expression is inverted to create an implicit reward for integration within the BT model, IPO instead attempts to approximate this equivalence relation by replacing the unknown $\pi_{\tiny \mbox{IPO}}(y|x)$ with some $\pi_\theta(y|x)$.  Although technically $r_{\tiny \mbox{IPO}}$ is also unknown, given samples $\{y_w,y_l,x\} \sim \calD_{\tiny \mbox{tr}}$, it is nicely shown in \cite{azar2024general} that ~~~$\ell_{\tiny \mbox{IPO}}(\pi_\theta,\pi_{\tiny \mbox{ref}},\lambda) ~~:=$
\begin{eqnarray} \label{eq:ipo_loss}  
 &&   \mathbb{E}_{\{y_1,y_2\} \sim \pi_{\tiny \mbox{ref}}(y|x), x\sim \calD} \left[ \left( \log \left[ \frac{\pi_\theta(y_1|x) \pi_{\tiny \mbox{ref}}(y_2|x)}{\pi_\theta(y_2|x) \pi_{\tiny \mbox{ref}}(y_1|x)}\right] - \frac{1}{\lambda}\big[ r_{\tiny \mbox{IPO}}(y_1,x) - r_{\tiny \mbox{IPO}}(y_2,x) \big] \right)^2 \right] \nonumber \\
    && = ~~~ \mathbb{E}_{\{y_w,y_l,x\} \sim \calD_{\tiny \mbox{tr}}} \left[ \left( \log \left[ \frac{\pi_\theta(y_w|x) \pi_{\tiny \mbox{ref}}(y_l|x)} {\pi_\theta(y_l|x) \pi_{\tiny \mbox{ref}}(y_w|x)}\right] - \frac{1}{2\lambda} \right)^2 \right]
\end{eqnarray}
provided $\calD_{tr}$ follows from (\ref{eq:preference_sampling}).  Note that this closed-form consistency is a direct consequence of how $r_{\tiny \mbox{IPO}}$ is defined in (\ref{eq:ipo_reward}) and will not generally hold for \textit{other} choices of the reward function.  Regardless, it is straightforward to minimize $\ell_{\tiny \mbox{IPO}}(\pi_\theta,\pi_{\tiny \mbox{ref}},\lambda)$ in its present form via SGD as with DPO.

% defined asymptotically in the limit of infinite training samples

\subsection{Flexible Quasi-Convex Generalizations} \label{sec:QPO_loss}

From the expressions above, it is clear that both DPO and IPO reduce to functions of $\log \left[ \frac{\pi_\theta(y_w|x) \pi_{\tiny \mbox{ref}}(y_l|x)} {\pi_\theta(y_l|x) \pi_{\tiny \mbox{ref}}(y_w|x)}\right]$ and a tunable hyperparameter $\lambda$.  As such, it is natural to consider extensions to broader choices in the form
\vspace*{-0.2cm}
\begin{equation} \label{eq:qpo_loss}
    \ell_{\tiny \mbox{QPO}}(\pi_\theta,\pi_{\tiny \mbox{ref}},\psi,\mu,\lambda) := \mathbb{E}_{\{y_w,y_l,x\} \sim \calD_{\tiny \mbox{tr}}} ~ \psi \left(\mu\left[ \frac{\pi_\theta(y_w|x) } {\pi_{\tiny \mbox{ref}}(y_w|x)} \right]- \mu\left[ \frac{\pi_\theta(y_l|x) } {\pi_{\tiny \mbox{ref}}(y_l|x)} \right], \lambda \right), 
\end{equation}
where $\mu : \mathbb{R}^+ \rightarrow \mathbb{R}$ is a monotonically increasing function (which generalizes the logarithm), and the function $\psi : \mathbb{R} \times \mathbb{R}^+ \rightarrow \mathbb{R}$ influences the overall loss shape.  We stipulate that $\psi$ is a differentiable \textit{quasi-convex} function \cite{greenberg1971review}; hence the chosen loss notation $\ell_{\tiny \mbox{QPO}}$ for quasi-convex preference optimization.  By definition of quasi-convexity, $\psi$ monotonically increases to the right or left away from the minimum.  

These specifications cover DPO and IPO as representative special cases, and include essentially all reasonable choices for a loss within this family, e.g., it is nonsensical to include multi-modal losses.  The generalized preference optimization (GPO) \cite{tang2024generalized} and \textit{f}-DPO \cite{wang2024beyond} frameworks are also special cases of QPO as defined herein.  With GPO, $\mu$ is a logarithm and $\psi$ is chosen as an arbitrary convex function (such as used by SLiC \cite{zhao2023slic}). Meanwhile \textit{f}-DPO involves $\psi(\cdot,\lambda) = -\log \sigma[\lambda( \cdot )]$ analogous to DPO but with $\mu = f'$, where $f'$ denotes the derivative of an $f$-divergence \cite{rubenstein2019practical}; given that $f$ must be convex, its derivative will necessarily be monotonically increasing.  In this way, the RLHF objective from (\ref{eq:rlhf_loss}) is still optimized via $f$-DPO, but with an  $f$-divergence replacing the KL term.

While overall quite general, we will nonetheless later demonstrate that any loss in the form of (\ref{eq:qpo_loss}) will unavoidably be saddled with certain limitations.  See also Appendix \ref{sec:related_work} for additional context w.r.t.~very recent and/or concurrent DPO enhancements that lie outside the scope of our present work.

\section{Comparative Analysis of Existing Approaches} \label{sec:analysis}

We now turn to comparative analysis of existing approaches, which all have ties relating back to the BT preference model.  Throughout this section we say that a policy $\pi^*$ is \textit{BT-optimal} at prompt $x$ if $p^*(y_1 \succ y_2|x)$ implies that $\pi^*(y_1|x) > \pi^*(y_2|x)$ for all response pairs $\{y_1,y_2\}$ with nonzero probability (as determined by the reference policy generating the preference data). Appendix \ref{sec:max_reward_derivation} introduces how $\pi^*$ can be formed.

% \tong{for all aforementioned methods, do they use BT model? If so, do we need to slightly emphasize that this is the typical choice?}

\subsection{Selective Preservation of Optimal Policies} \label{sec:equiv_criteria_analysis}

% $\pi^{*}$ is an optimal policy w.r.t.~any arbitrary measure of human preferences and 

Consider the following plausible scenario, variations of which are likely to occur (at least in varying degrees) with real-world data. Suppose the support of prompts generated by $\calD_x$ partitions as $d_x^{good} \cup d_x^{bad}$, with $d_x^{good} \cap d_x^{bad} = \emptyset$.  Furthermore, assume we have access to a reference policy $\pi_{\tiny \mbox{ref}}$ such that $\pi_{\tiny \mbox{ref}} = \pi^{*}$ for $x \in d_x^{good}$ and $\mbox{dist}[\pi_{\tiny \mbox{ref}},~\pi^{*}] \gg 0$ for $x \in d_x^{bad}$, where $\mbox{dist}[\cdot, \cdot]$ is an arbitrary distance measure.   In other words, when evaluated w.r.t.~a policy $\pi^{*}$ that proportionally reflects human preferences, $\pi_{\tiny \mbox{ref}}$ performs ideally on a subset of prompts but not on others.

This dichotomy provides a useful lens for examining certain loss function properties.  In particular, we would like any policy that minimizes a candidate loss to preserve $\pi_{\tiny \mbox{ref}}$ for prompts $x \in d_x^{good}$, while pushing away from $\pi_{\tiny \mbox{ref}}$ towards $\pi^*$ for prompts $x \in d_x^{bad}$.  However, because of uniform regularization effects intrinsic to the QPO loss, it is not actually possible to achieve even this modest objective.  

\begin{theorem} \label{thm:qpo_equivalence}
(Informal version)~~Given the prompt partitioning, reference policy, and optimal policy described above, define $\hat{\pi}_\theta^{\tiny \mbox{QPO}} := \arg\min_{\pi_\theta} \ell_{\tiny \mbox{QPO}}(\pi_\theta,\pi_{\tiny \mbox{ref}},\psi,\lambda)$ for any fixed selection of $(\psi,\lambda)$.  Then under relatively mild assumptions on the labeled responses in $\calD_{tr}$, if $\mbox{dist}[\hat{\pi}_\theta^{\tiny \mbox{QPO}},~\pi^{*}] <  \mbox{dist}[\pi_{\tiny \mbox{ref}},~\pi^{*}]$ for $x \in d_x^{bad}$, then $\mbox{dist}[\hat{\pi}_\theta^{\tiny \mbox{QPO}},~\pi^{*}] > 0$~ for $x \in d_x^{good}$.
\end{theorem}

The proof and formal version are provided in Appendix \ref{app:formal_version}, while Figure \ref{fig:desiderata_fig}(\textit{left}) below provides an illustration. The somewhat unexpected implication here is that if we minimize any possible QPO loss in the form  of (\ref{eq:qpo_loss}) and improve the policy quality in areas where $\pi_{\tiny \mbox{ref}}$ performs poorly w.r.t.~$\pi^*$, then it \textit{must also be the case that performance  becomes worse in areas where} $\pi_{\tiny \mbox{ref}}$ \textit{was originally optimal}.  This phenomena represents an unavoidable trade-off when we restrict ourselves to using a QPO loss, of which DPO and IPO (as well as GPO and $f$-DPO) are special cases inheriting the same limitation.  The core issue here is that QPO losses \textit{unselectively} apply the same regularization, starting from the same initialization point, to both good and bad cases relative to $\pi^*$. % please see Appendix \ref{app:qpc_regularization} for an illustrative example of this phenomena.  

% \tong{Can we consider to move this interpretation earlier? I think this is understandable in pure language, and with this in mind theorem 1 can be easier to understand.}\david{What about even up to the intro section, first bullet point? Or did you mean earlier in this section?}

\subsection{Interpolation Capabilities} \label{sec:interpolate_criteria_analysis}

As the underlying goal shared by all approaches is to \textit{balance} proximity to a reference policy $\pi_{\tiny \mbox{ref}}$ with respect for the human preference model $p^*$, a non-negative trade-off parameter $\lambda \in [a,b]$ that allows for interpolating between these competing objectives is inevitable, where $a \in \mathbb{R}$ and $b \in \mathbb{R}$ are lower and upper bounds respectively.\footnote{Depending on the method, if $a = 0$ or $b = \infty$ we may replace the $\lambda$ range with an open set.}  In this section we examine more closely the nature of loss function minimizers as $\lambda$ is varied, zooming in on their behavior in the limit as $\lambda \rightarrow a$ and $\lambda \rightarrow b$.  To this end, we first introduce the following definitions :

% \david{TODO: Add references to general regularized losses with trade-off parameters.}

% w.r.t.~an arbitrary preference optimization loss $\ell(\pi_\theta,\pi_{\tiny \mbox{ref}},\lambda)$ and $\lambda \in [a,b]$

% vis-\`a-vis these objectives approaches either 

\begin{definition}
We say that an arbitrary preference optimization loss $\ell(\pi_\theta,\pi_{\tiny \mbox{ref}},\lambda)$ satisfies the \textbf{strong interpolation criteria (SIC)} if the following conditions hold:
\begin{enumerate}
    \item $\lim_{\lambda \rightarrow a} \arg\min_{\pi_\theta} \ell(\pi_\theta,\pi_{\tiny \mbox{ref}}, \lambda) = \pi^*$; 
    
    \item $\lim_{\lambda \rightarrow b} \arg\min_{\pi_\theta} \ell(\pi_\theta,\pi_{\tiny \mbox{ref}}, \lambda) = \pi_{\tiny \mbox{ref}}$;

    \item For all other $\lambda \in (a,b)$, the optimal policy interpolates between the above two extremes.
\end{enumerate}
\end{definition}

    % \item With regard to the \textit{lower} bound on $\lambda$, sample pairs $(y,y')$ drawn from the optimal policy $\pi_\theta^a := \arg\min_{\pi_\theta} \ell(\pi_\theta,\pi_{\tiny \mbox{ref}}, a)$ are such that the probability that $y$ is preferred over $y'$ equals $p^*(y \succ y'|x)$, i.e., the ground-truth BT preference model (or at least the empirical version of it for finite samples) is recovered when $\lambda = a$.

\begin{definition}
For any prompt $x$ and response $y$ define\footnote{See Appendix \ref{sec:max_reward_derivation} for the derivation of the right-hand equality in (\ref{eq:max_reward_derivation}).}  
\begin{equation} \label{eq:max_reward_derivation}
    \pi^\delta(y|x) ~~:=~~ \arg\max_{\pi_\theta} \mathbb{E}_{y \sim \pi_{\theta}(y|x)}\big[ r^*(y,x) \big] ~~=~~ \left\{ \begin{array}{cc}
      1   & \mbox{if}~~ y = \arg\max_{y'} \pi^*(y|x) \\
      0   & \mbox{otherwise.}
    \end{array} \right.
\end{equation}
In this way, $\pi^\delta(y|x)$ assigns probability one to the mode of $\pi^*(y|x)$, i.e., akin to a delta function with no generation diversity.  We then say that a loss $\ell(\pi_\theta,\pi_{\tiny \mbox{ref}},\lambda)$ satisfies the \textbf{weak interpolation criteria (WIC)} analogously to the SIC, only  for the lower bound we instead require ~~ $\lim_{\lambda \rightarrow a} \arg\min_{\pi_\theta} \ell(\pi_\theta,\pi_{\tiny \mbox{ref}}, \lambda) = \pi^\delta$.
\end{definition}

In summary, the only difference between these interpolation criteria is their limiting behavior w.r.t.~the lower bounding $\lambda$; for the SIC we approach the BT-optimal policy, while for the WIC we approach a degenerate policy with all probability mass restricted to the \textit{mode} of the BT-optimal policy.  We remark that both the SIC and WIC cannot be simultaneously satisfied unless $\pi^*$ itself is a degenerate delta function. 
We now explore how these distinctions are reflected in the behavior of DPO and IPO loss minimizers, with Figure \ref{fig:desiderata_fig}(\textit{middle}) illustrating the basic concepts.
%\tong{Does the analysis hold for generalized *-POs?-- I see theorem 2 analyzes QPO, maybe we also mention here.}\david{I think maybe for $f$-DPO it holds trivially.  But need to double-check this.}

% (i.e., the probability assignments are not exactly equal to zero or one)
\begin{proposition} \label{prop:DPO_interpolation}
    Assume preference data distributed according to $\calD_{tr}$ from (\ref{eq:preference_sampling}), and that $p^*(y_1 \succ y_2 |x) \in (0,1)$ for all responses with $\pi_{\tiny \mbox{ref}}(y|x) > 0$.  Then the DPO loss from (\ref{eq:dpo_loss}) satisfies the WIC (but not the SIC).
\end{proposition}
In terms of practical applicability of this result, there exists one important caveat: the empirical distribution of a finite set of labeled preference data need not actually satisfy the conditions of Proposition \ref{prop:DPO_interpolation}.  For example, suppose for each prompt $x \in \calD_x$ we collect only two responses $\{y_1,y_2\}$ along with a single preference label $z$, which together produce the tuple $\{y_w,y_l,x\}$.  In this scenario, which reflects certain publicly-available human preference datasets \cite{bai2022training,ganguli2022red}, the empirical distribution of preferences will be $p^*(y_w \succ y_l | x) = 1 \notin (0,1)$ for all $x \in \calD_x$.  Notably, Proposition \ref{prop:DPO_interpolation} will \textit{not} hold, and in particular, it can be easily shown that minimizers of any valid $f$-DPO loss will be \textit{completely independent} of $\pi_{\tiny \mbox{ref}}$ for all $\lambda \in (0,\infty)$; in other words, \textit{no interpolation occurs at all}; see Appendix \ref{sec:f-DPO_analysis} for the derivation. A similar observation specific to DPO (but not $f$-DPO) can be found in \cite{ahmadian2024back}.  The fact that DPO-based solutions may still reflect $\pi_{\tiny \mbox{ref}}$ in practice, and more-so as $\lambda$ increases, relates to implicit constraints and subtle regularization effects as discussed further in Section \ref{sec:constraints} and Appendix \ref{sec:dpo_noise_adaptive_loss}.

% is an artifact of implicit constraints/regularization imposed during training.

% The fact that DPO can still at times operate effectively in such situations, with some dependency on $\pi_{\tiny \mbox{ref}}$ is due to implicit constraints imposed during the training process, a topic we will revisit in Section \ref{sec:constraints}.

\begin{proposition} \label{prop:IPO_interpolation}
Assume preference data distributed according to $\calD_{tr}$ from (\ref{eq:preference_sampling}).  Then the IPO loss from (\ref{eq:ipo_loss}) satisfies the WIC (but not the SIC).
\end{proposition}
Comparing Proposition \ref{prop:IPO_interpolation} with Proposition \ref{prop:DPO_interpolation}, we observe that IPO maintains its ability to interpolate under broader conditions than DPO, particularly in the empirical sampling regime involving binary probability values.  That being said, neither DPO nor IPO satisfy the SIC, which motivates consideration of alternative losses that do, at least if our priority is to actually achieve the SIC (which of course may depend on the application scenario).  For this purpose, it turns out that selections \textit{beyond} the family of QPO objectives (which includes DPO, $f$-DPO, and IPO) are necessary per the following:
\begin{theorem} \label{thm:qpo_interpolation_limitation}
Assume preference data distributed according to $\calD_{tr}$ from (\ref{eq:preference_sampling}).  Then no possible QPO loss from (\ref{eq:qpo_loss}) will satisfy the SIC.
\end{theorem}
Section \ref{sec:new_objectives} will consider objectives outside of the QPO family which circumvent this limitation.

\subsection{Impact of Optimization Constraints} \label{sec:constraints}

Originally in \cite{rafailov2024direct}, and later supported by follow-up analysis \cite{azar2024general}, it has been shown that minimizing the DPO loss $\ell_{\tiny \mbox{DPO}}(\pi_\theta, \pi_{\tiny \mbox{ref}},\lambda)$ is effectively the same as minimizing the RLHF loss $\ell_{\tiny \mbox{RLHF}}\left(\pi_\theta, \pi_{\tiny \mbox{ref}}, r^*, \lambda \right)$ with optimal reward model $r^*$.  But there is a pivotal assumption underlying this association which previous analysis has not rigorously accounted for.  Specifically, the key equalities that facilitate the DPO and IPO reparameterizations, namely (\ref{eq:closed_form_rlhf_reward}) and (\ref{eq:ipo_equivalence_relation}) (and the analogous for $f$-DPO), are all predicated on the solution of an \textit{uncononstrained} optimization problem over an arbitrary policy $\pi_\theta$.  

\begin{figure*}[t]
    %\centering
    \hspace*{-1.0cm} \includegraphics[width=16cm]{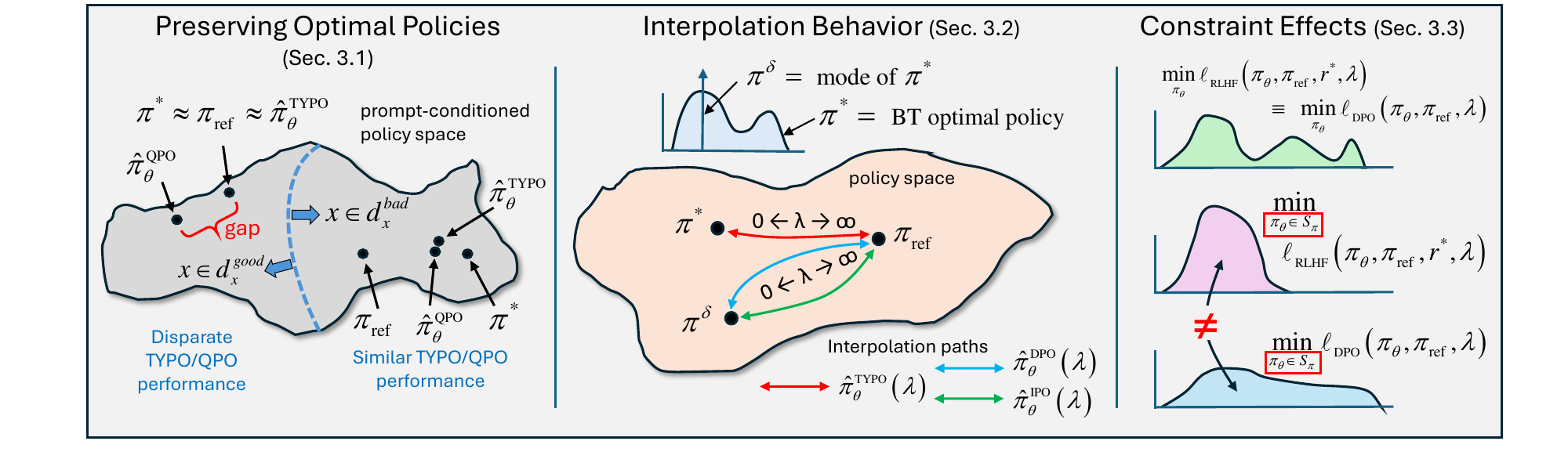}
    \vspace*{-0.7cm}
    \caption{Desiderata visualizations, including added context w.r.t.~our proposed TYPO approach.}
    \label{fig:desiderata_fig}
    %\vspace*{-0.8cm}
\end{figure*}

However, when actually training models in real-world settings, constraints will always exist, whether implicitly or explicitly.  Such constraints stem from any number of factors including the model architecture/capacity limitations, early stopping, weight decay, drop-out regularization, machine precision, and so on.  Hence in reality we are never exactly minimizing some preference loss $\ell\left(\pi_\theta, \pi_{\tiny \mbox{ref}}, \lambda \right)$ over any possible $\pi_\theta$ (as assumed by DPO, IPO, and $f$-DPO derivations).  Instead, we must consider properties of the \textit{constrained} problem  $\min_{\pi_\theta \in \calS_\pi} \ell\left(\pi_\theta, \pi_{\tiny \mbox{ref}}, \lambda \right)$, where $\calS_\pi$ is a constraint set.  For example, if we restrict training to a single epoch with a fixed learning rate, then $\calS_\pi$ can be viewed as the set of all points reachable within a limited number of SGD updates. 
% \tong{Is there a straightforward way to put it? Maybe something like ``there exist $\calS_\pi$ as a constraint set such that blah blah''.}\david{Yeah it is tricky how to word this (but I modified a bit).  One issue is risking the perception of there just being some weird hacked constraints to make it fail, but it is otherwise mostly true (which is not accurate I think).}
\begin{theorem} \label{thm:dpo_constraints}
Let $\calS_\pi$ denote a constraint set on the learnable policy $\pi_\theta$.  Then we can have that
    \begin{equation} \label{eq:dpo_constraint_equivalence}
 \arg\min_{\pi_\theta \in \calS_\pi}   \ell_{\tiny \mbox{RLHF}}\left(\pi_\theta, \pi_{\tiny \mbox{ref}}, r^*, \lambda \right) ~ \neq ~ \arg\min_{\pi_\theta  \in \calS_\pi} \ell_{\tiny \mbox{DPO}}(\pi_\theta, \pi_{\tiny \mbox{ref}},\lambda).
\end{equation}
\end{theorem}
\vspace*{-0.2cm}
As can be observed by the proof in Appendix \ref{sec:dpo_contraint_proof}, the difference between the two is akin to the difference between applying a constraint to a trainable policy with respect to either the forward or backward KL divergence, which are generally quite distinct \cite{Bishop06}; see also Figure \ref{fig:desiderata_fig}(\textit{right}).  There are several important consequences of this result worth considering:
\begin{itemize}
    \item As discussed in Section \ref{sec:interpolate_criteria_analysis}, the DPO-based losses can have degenerate unconstrained minimizers that completely ignore $\pi_{\tiny \mbox{ref}}$ on certain real-world datasets; therefore counter-measures like early stopping are imposed that effectively introduce a $\calS_\pi$ that dramatically alters the estimated policy.  But in doing so, the inequality from (\ref{eq:dpo_constraint_equivalence}) is introduced and so \textit{we can no longer say that DPO provides an optimal implicit reward for the original RLHF problem}, i.e., the original connection is now ambiguous.  
    
    \item As such, the value of DPO in practice (and indeed it often does work well) cannot be unreservedly attributed to its original affiliation with an optimal RLHF solution, and instead, should be evaluated based on properties of $\min_{\pi_\theta  \in \calS_\pi} \ell_{\tiny \mbox{DPO}}(\pi_\theta, \pi_{\tiny \mbox{ref}},\lambda)$.  See Appendix \ref{sec:dpo_noise_adaptive_loss} for one step in this direction.
    
    \item To further illustrate the above points, in Appendix \ref{sec:gaussian_dpo} we rederive the DPO loss from scratch \textit{based solely on a Gaussian estimation perspective that is completely unrelated to RLHF}.  But of course we do not actually believe that binary human preference data are really Gaussian.  Instead, this exercise serves to highlight that what matters \textit{are properties of the underlying loss when deployed in practice}, not necessarily the assumptions made in deriving the loss in the first place.
    \item Other losses based on unconstrained RLHF-based reparameterizations in the $f$-DPO and IPO families may be similarly influenced by the inevitable introduction of policy constraints.
\end{itemize}

% \david{TODO: Add why optimization constraints cannot be ignored to motivate the significance of this result.  Also, may be preferable to move this section earlier.}

% \david{Mention Gaussian derivation of DPO to motivate that only the final loss matters, not the original steps that get us there.}

% \david{Possibly extend to IPO, $f$-DPO as well; mention that some GPO approaches are not based on any such reparameterization though.}

% \david{Possibly introduce augmentation criteria in new subsection here.}

% .  Superscripts on policies denote which loss produced them.

\vspace*{-0.2cm}
\section{New Objectives for Human Preference Optimization} \label{sec:new_objectives}
\vspace*{-0.2cm}

Motivated by the analysis in Section \ref{sec:analysis} and illustrated in Figure \ref{fig:desiderata_fig}, we next examine alternative objective functions adhering to the following desiderata:
\begin{enumerate}
    \item \textbf{Perservation:}~~Capable of selectively preserving an optimal policy in ideal regimes, while \textit{simultaneously} improving the policy in regions of poor performance (from Section \ref{sec:equiv_criteria_analysis}); 
    \item \textbf{Interpolation:}~~Smoothly interpolates between the BT-optimal policy and the reference policy, i.e., it achieves the SIC (from Section \ref{sec:interpolate_criteria_analysis});
    \item \textbf{Constraints:}~~Independent of any derivation or required equivalence/reparameterization that no longer holds upon the introduction of constraints (from Section \ref{sec:constraints}).
\end{enumerate}
We label the our new objective  $\ell_{\tiny \mbox{TYPO}}$ to highlight the potential ability to ``\textit{\underline{t}ame \underline{y}our \underline{p}reference \underline{o}ptimization}'' (and ``\textit{lower typos}'') by explicitly targeting these desiderata.

% We break $\ell_{\tiny \mbox{TYPO}}$ into two categories, compositional losses $\ell^c_{\tiny \mbox{TYPO}}$ and regression-based losses $\ell^r_{\tiny \mbox{TYPO}}$, analyzing each in turn below.

\subsection{TYPO Objective Function}
% \subsection{Compositional Approach}

Consider a loss, composed of separable supervised and unsupervised factors, in the general form
\begin{eqnarray} \label{eq:DPO_reg_general}
&& \hspace*{-0.6cm} \ell_{\tiny \mbox{TYPO}}(\pi_\theta,\pi_{\tiny \mbox{ref}},\lambda) ~~  := ~~ \ell_{\tiny \mbox{sup}}(\pi_\theta)  ~+~ \lambda \ell_{\tiny \mbox{unsup}}(\pi_\theta,\pi_{\tiny \mbox{ref}}) ~~ =\\
&& \mathbb{E}_{\{y_w,y_l,x\} \sim \calD_{tr}} \Big[  d_{\tiny \mbox{sup}} \big[  \pi_\theta(y_w | x), \pi_\theta(y_l | x)  \big] \Big] ~+~ \lambda \mathbb{E}_{y \sim \pi_{\tiny \mbox{ref}}(y|x), x\sim \calD_x} \Big[ d_{\tiny \mbox{unsup}}\big[\pi_\theta(y|x), \pi_{\tiny \mbox{ref}}(y|x) , \big] \Big], \nonumber
\end{eqnarray}
where $d_{\tiny \mbox{sup}}$ serves as a supervised penalty over labeled training tuples $(x,y_w,y_l)$ while $d_{\tiny \mbox{unsup}}$ represents an additional regularization term independent of labeled preferences. We remark that objectives in the form of (\ref{eq:DPO_reg_general}) are natural candidates for SGD given that all sampling is independent of $\theta$, unlike the typical regularized loss adopted by RLHF, which requires samples from $\pi_\theta(y|x)$.  

% We next introduce specific choices for the supervised an unsupervised factors designed to reflect the desiderata from Section \ref{sec:desiderata}.

\paragraph{Supervised Term:}   After first defining 
\vspace*{-0.4cm}
\begin{equation} \label{eq:theta_preference_prob}
   p_\theta(z| y_1, y_2,x) := \left\{ \begin{array}{cc}
       \frac{\pi_\theta(y_1|x)}{\pi_\theta(y_1|x)+\pi_\theta(y_2|x)} & \mbox{if } z = 1\\
        \frac{\pi_\theta(y_2|x)}{\pi_\theta(y_1|x)+\pi_\theta(y_2|x)} & \mbox{if } z = 0
   \end{array}   \right. 
\end{equation}
\vspace*{-0.1cm}
we then consider the supervised term
\begin{eqnarray} \label{eq:new_pen_1}
\ell_{\tiny \mbox{sup}}(\pi_\theta) & = &  \mathbb{E}_{\{y_1,y_2\} \sim \pi_{\tiny \mbox{ref}}(y|x),x \sim \calD_x} \Big[  \mathbb{KL}\big[ p^*(z| y_1, y_2,x) || p_\theta(z| y_1, y_2,x) \big] \Big] \nonumber  \\
& \equiv & \mathbb{E}_{\{y_w,y_l,x\} \sim \calD_{tr}} \left[  \log\left(1 +  \frac{\pi_\theta(y_l | x)}{\pi_\theta(y_w | x)} \right)  \right].
\end{eqnarray}
Please see Appendix \ref{sec:simplified_penalty} for the derivation of this equivalence.  Importantly here, because the KL-divergence is minimized iff $p^*(z| y_1, y_2,x) = p_\theta(z| y_1, y_2,x)$, unlike an arbitrary reward, the optimal solution to $\ell_{\tiny \mbox{sup}}(\pi_\theta)$ will \textit{necessarily} recover the BT-optimal  distribution as will be analyzed  below.

\paragraph{Unsupervised Term:}  For the  unsupervised term in (\ref{eq:DPO_reg_general}) we simply adopt
\vspace*{-0.1cm}
\begin{equation} \label{eq:standard_KL}
\ell_{\tiny \mbox{unsup}}(\pi_\theta,\pi_{\tiny \mbox{ref}}) = \mathbb{E}_{ x\sim \calD_x} \Big[\mathbb{KL}\big[\pi_{\tiny \mbox{ref}}(y|x) || \pi_\theta(y|x) \big] \Big] \equiv -\mathbb{E}_{y \sim \pi_{\tiny \mbox{ref}}(y|x), x\sim \calD_x} \Big[\log \pi_\theta(y|x) \Big],
\end{equation}
ignoring terms independent of $\pi_\theta$.  Like (\ref{eq:new_pen_1}), this expression also does not require sampling from $\pi_\theta$.  That being said, (\ref{eq:standard_KL}) can exploit out-of-preference data (meaning unlabeled responses), and prior work \cite{li2023policy} has argued for the merits of using such data in broader RLHF contexts.  (It may also be reasonable to consider switching $\ell_{\tiny \mbox{unsup}}(\pi_\theta,\pi_{\tiny \mbox{ref}})$ to a reverse-KL term and optimize with REINFORCE per general observations from \cite{ahmadian2024back}; however, we do not pursue this direction further here.)

\vspace*{-0.2cm}
\subsection{$\ell_{\tiny \mbox{TYPO}}$ Properties} \label{sec:typo_properties}
\vspace*{-0.2cm}

Notable attributes of $\ell{\tiny \mbox{TYPO}}(\pi_\theta,\pi_{\tiny \mbox{ref}},\lambda)$ w.r.t.~the three desiderata from above are as follows: 

\begin{proposition} \label{prop:TYPO_preservation}
    Under the same setup as Theorem \ref{thm:qpo_equivalence}, let $\hat{\pi}_\theta^{\tiny \mbox{TYPO}} := \arg\min_{\pi_\theta} \ell_{\tiny \mbox{TYPO}}(\pi_\theta,\pi_{\tiny \mbox{ref}},\lambda)$, instantiated using (\ref{eq:new_pen_1}) and (\ref{eq:standard_KL}). Then $\hat{\pi}_\theta^{\tiny \mbox{TYPO}} = \pi^{*}$~ for all $x \in d_x^{good}$ including in cases where $\mbox{dist}[\hat{\pi}_\theta^{\tiny \mbox{TYPO}},~\pi^{*}] <  \mbox{dist}[\pi_{\tiny \mbox{ref}},~\pi^{*}]$ for $x \in d_x^{bad}$.
\end{proposition}
Per this result, minimizers of $\ell_{\tiny\mbox{TYPO}}(\pi_\theta,\pi_{\tiny \mbox{ref}},\lambda)$ are capable of preserving $\pi_{\tiny \mbox{ref}}$ in regions $d_x^{good}$ where performance is strong relative to $\pi^*$, while concurrently improving performance in other areas where it is not.  Figure \ref{fig:desiderata_fig}(\textit{left}) visualizes this unique TYPO capability.

\begin{proposition}  \label{prop:TYPO_interpolation}
The loss $\ell{\tiny \mbox{TYPO}}(\pi_\theta,\pi_{\tiny \mbox{ref}},\lambda)$, when instantiated using (\ref{eq:new_pen_1}) and (\ref{eq:standard_KL}), satisfies the SIC.
\end{proposition}

Figure \ref{fig:desiderata_fig}(\textit{middle}) contrasts this property with the WIC achieved by prior methods.  We also remark that none of the derivations used to motivate $\ell{\tiny \mbox{TYPO}}(\pi_\theta,\pi_{\tiny \mbox{ref}},\lambda)$ rely on unconstrained optimization to form a reparameterized objective function as with DPO, $f$-DPO, and IPO.  As such, the inevitable introduction of such constraints in practice does not compromise the TYPO origin story. In other words, since TYPO is not based on any implicit association with RLHF in the first place, adding constraints that might otherwise compromise such an association pose no issue.

% \tong{I'd assume reviewers would question how about the relationship between $\ell{\tiny \mbox{TYPO}}$ and $\ell{\tiny \mbox{RLHF}}$.}\david{Yes they might.  It correlates with the relationship between TYPO and DPO, but there is some nuance relating to constraints and the training data distribution.  We can discuss if needed.}

% Note that per the discussion in Section \ref{sec:formal_criteria_analysis}, it is not possible to simultaneously satisfy the SIC and WIC, so in terms of the formal criteria, this result is optimal.  With regard to the informal criteria from Section \ref{sec:informal}, both $\ell^c_{\tiny \mbox{TYPO}}$ and $\ell^r_{\tiny \mbox{TYPO}}$ satisfy all three. \david{TODO: Add details why this is true.}

\vspace*{-0.2cm}
\section{Empirical Validation} \label{sec:experiments}
\vspace*{-0.2cm}
Although more of an analysis-driven contribution, our core insights from Sections \ref{sec:analysis} and \ref{sec:new_objectives} can nonetheless benefit from empirical corroboration.  To this end, we first present a series of experiments adapted from \cite{azar2024general} to highlight aspects of TYPO behavior vis-\`a-vis our proposed desiderata.  As the most relevant published points of reference, we contrast with DPO, IPO, and $f$-DPO; for the latter we choose the Jensen–Shannon divergence, which next to the reverse-KL implicitly assumed by DPO, performed well in prior experiments \cite{wang2024beyond}.  Later we test using the Anthropic Helpfulness and Harmlessness (HH) real-world preference dataset \cite{bai2022training, ganguli2022red}.  For space considerations, some experiment details, including hyperparameters and training setups, are deferred to Appendix~\ref{appendix:exp_detail}.

% ... compare with recent published methods. 

% As our contribution has more of an analytical bent, our experiments largely focus on corroborating various aspects

\begin{figure}
    \centering
    \includegraphics[width=0.8\textwidth]{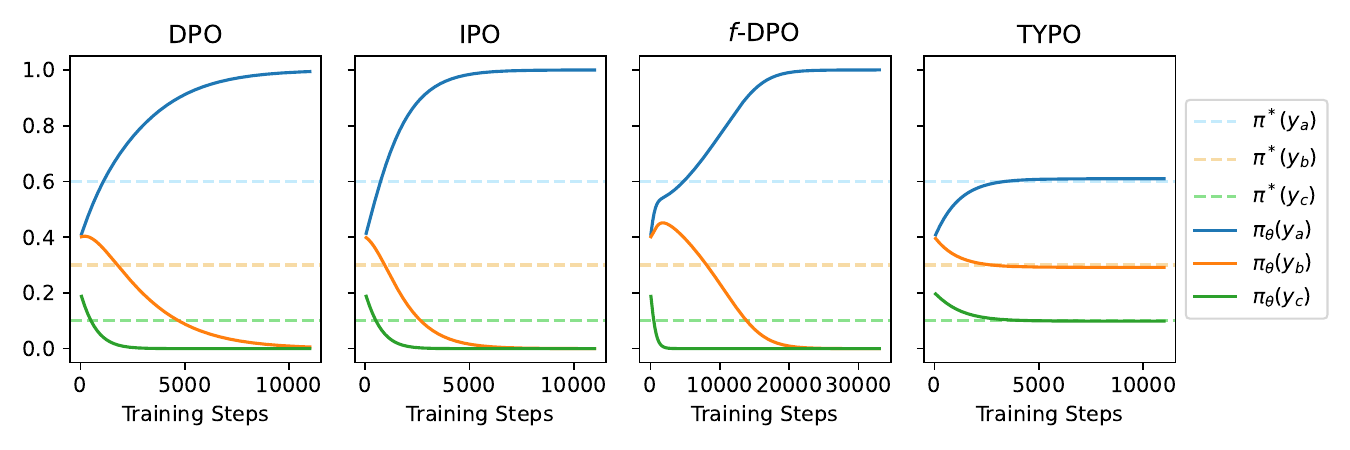}
    \vspace*{-0.2cm}
     \caption{\textit{Support for Sections \ref{sec:interpolate_criteria_analysis} and \ref{sec:typo_properties} interpolation analysis.} Dashed lines represent BT-optimal preference probabilities $\pi^*$, while solid lines are model learning curves for $\lambda = 10^{-5}$ (small).  Only TYPO converges to $\pi^*$, others converge to $\pi^\delta$.}
    \label{fig:exp_interpolation_curves_small_lambda}
\end{figure}

\paragraph{Interpolation Tests:}  As in \cite{azar2024general} we consider the bandit setting with a discrete space of three responses/actions $\calY = \{y_a,y_b,y_c\}$ and create a dataset of labeled pairs as $\big\{\{y_a,y_b\}, \{y_b,y_c\}, \{y_a, y_c \}  \big\}$, i.e., a total ordering consistent with the BT model. Preferences are assigned via $p(y_1 \succ y_2)$ computed using (\ref{eq:BT_optimal_model_specs}) with $\pi^*(y_a) = 0.6$, $\pi^*(y_b) = 0.3$, and $\pi^*(y_c) = 0.1$.  Furthermore, again following \cite{azar2024general} we form our trainable policy as $\pi_\theta(y_i) = \mbox{softmax}[\theta_i]$ with $\theta \in \mathbb{R}^3$ optimized using Adam over each different preference loss.  Results using a small $\lambda = 10^{-5}$ are shown in Figure \ref{fig:exp_interpolation_curves_small_lambda}, where we observe that TYPO closely converges to the BT-optimal solution, while DPO and IPO converge to $\pi^\delta$ (the mode of $\pi^*$) consistent with Propositions \ref{prop:DPO_interpolation} (DPO), \ref{prop:IPO_interpolation} (IPO), and \ref{prop:TYPO_interpolation} (TYPO), as well as Theorem \ref{thm:qpo_interpolation_limitation} which applies to $f$-DPO.  Additional interpolation results traversing different $\lambda$ towards the upper limit are presented in Appendix \ref{app:additional_interpolation_results}.

\begin{figure} 
    \centering
    \vspace*{-0.3cm}
    \includegraphics[width=\textwidth]{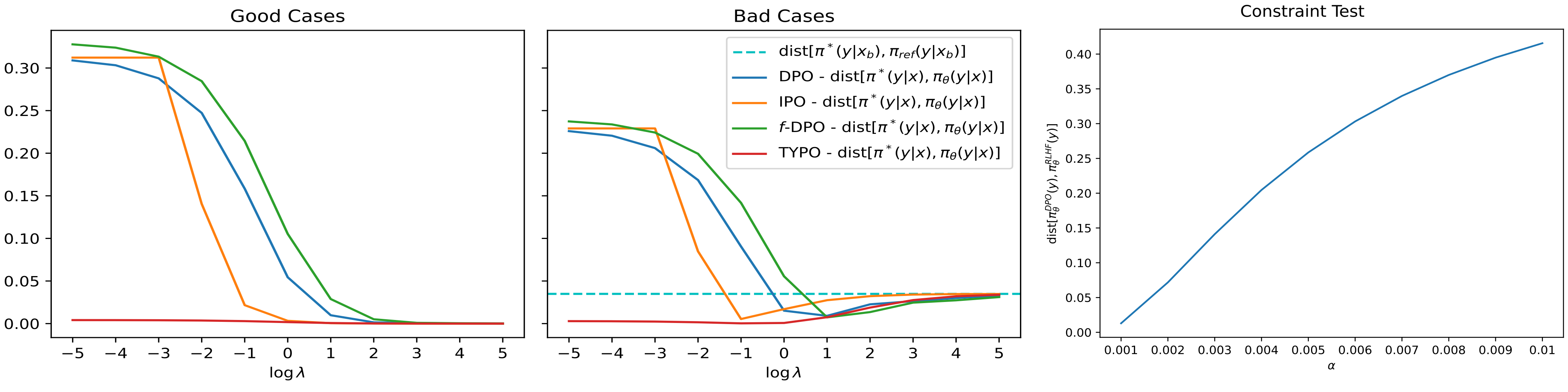}
    \vspace*{-0.6cm}
    \caption{\textit{Preservation tests} varying $\lambda$ (left and middle plots); unlike TYPO, existing approaches are unable to both retain negligible error on the good cases while improving performance (over the dashed line representing the reference model) on the bad cases. \textit{Constraint test} varying $\alpha$ and plotting $\mbox{dist}[\hat{\pi}_\theta^{\tiny \mbox{DPO}},~\hat{\pi}_\theta^{\tiny \mbox{RLHF}}]$ (right plot); DPO is no longer equivalent to RLHF with an optimal reward once an additional constraint/regularization factor is introduced.}
    \vspace*{-0.3cm}
    \label{fig:preservation_test}
\end{figure}

\paragraph{Preservation Tests:} We next modify the setting from above to include two input prompts $\{x_g, x_b\}$ chosen such that $x_g \in d_x^{good}$ and $x_b \in d_x^{bad}$ sampled with equal probability.  We then specify the corresponding response space $\calY(x_g) = \{y_{ga},y_{gb},y_{gc}\}; ~~~ \calY(x_b) = \{y_{ba},y_{bb},y_{bc}\}$ and prompt-dependent probabilities (see Appendix \ref{appendix:exp_detail_interpolation}).  For the reference policy we set $\pi_{\tiny \mbox{ref}}(y|x_g) = \pi^*(y|x_g)$ and $\pi_{\tiny \mbox{ref}}(y|x_b) \neq \pi^*(y|x_b)$.  We generate pair-wise preference data as before, only now with prompt-dependent responses.  Results shown in Figure \ref{fig:preservation_test}(\textit{left} \& \textit{middle}) are in direct accordance with Theorem \ref{thm:qpo_equivalence} and Proposition \ref{prop:TYPO_preservation}, whereby TYPO is the only approach that preserves a strong policy with prompt $x_g \in d_x^{good}$ while at the same time improving performance relative to $\pi_{\tiny \mbox{ref}}$ for $x_b \in d_x^{bad}$ over all $\lambda$.

\paragraph{Constraint Tests:}  We probe the extent to which learning constraints can interfere with the equivalence between DPO and RLHF implemented with an optimal reward function.  To this end, we adopted the same data generation setup as in the interpolation experiments from above.  We then train policies to separately minimize the right- and left-hand sides of (\ref{eq:dpo_constraint_equivalence}), but with one key modification: we added an identical penalty function $\alpha \|\pi_\theta\|_2^2$ to both models to instantiate weight decay (a typical form of constraint used in practice), where $\alpha \geq 0$ is a tunable hyperparameter.  Figure \ref{fig:preservation_test}(\textit{right}) plots the distance ($y$-axis) between learned policies from RLHF and DPO as $\alpha$ is varied.  Consistent with the original DPO derivations and analysis from \cite{rafailov2024direct}, we observe negligible error when $\alpha = 0$ given that unconstrained DPO is explicitly designed to mimic RLHF with an optimal reward $r^*$.  However, in accordance with our Theorem \ref{thm:dpo_constraints}, as $\alpha > 0$ increases, the distance between RLHF and DPO grows considerably, and their relationship is no longer clear-cut.

% This refers to verifying Theorem \ref{thm:dpo_constraints} (and an extension for IPO if there is time).  For this I think we could reuse the same setup as for the interpolation tests, no need for multiple prompts.  Basically, the idea would be to minimize separately the right and left sides of equation (\ref{eq:dpo_constraint_equivalence}) and show that they are different once we add constraints.  And to make it easier, we could implement the constraints just by adding an extra regularization factor to both loss functions, for example, something like $\alpha \| \theta \|_2^2$ with $\alpha > 0$.  Then the final plot would look like this:  x-axis would be $\alpha$ and y-axis would be the gap between the right and left sides of equation (\ref{eq:dpo_constraint_equivalence}).  Anyway, we could save this one for last and just discuss on Monday if that is easier.

% \begin{figure}
%     \centering
%     \includegraphics[width=0.6\textwidth]{figures/toy_exp_constraint_1.pdf}
%     \caption{Caption}
%     \label{fig:enter-label}
% \end{figure}

\begin{wrapfigure}{r}{0.4\textwidth}
\vspace*{-0.8cm}
  \begin{center}
    \includegraphics[width=0.4\textwidth]{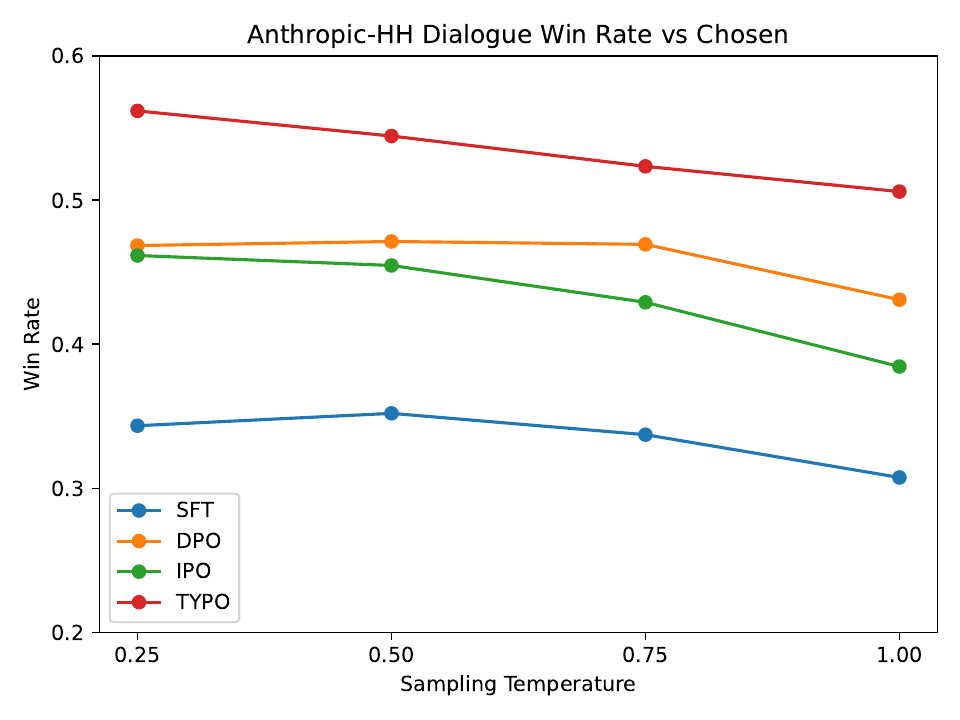}
  \end{center}
  \vspace*{-0.3cm}
  \caption{Real-world example.}
  \label{fig:hh_results}
\end{wrapfigure}

\paragraph{Testing on Anthropic HH Preference Data:}  Finally, to explore TYPO capabilities in a real-world scenario, we train a Pythia 2.8B model \cite{biderman2023pythia} on the Anthropic Helpfulness and Harmlessness (HH) preference dataset \cite{bai2022training, ganguli2022red} as previously used in \cite{rafailov2024direct}. Following their settings, we first execute supervised fine-tuning (SFT) on the Pythia model using $y_w$ values as the target response. We then use this SFT model as $\pi_{\tiny \mbox{ref}}$ for training DPO, IPO and TYPO.  Given that alignment results (our focus) from \cite{wang2024beyond} already show that reverse KL (i.e., DPO) works best among $f$-divergences, we do not compare with other $f$-DPO selections here. We use GPT-4 to evaluate the win rate of the generated responses from each model against the chosen $y_w$ on the test set for single turn dialogues.  We emphasize that our comparisons cover \textit{both} helpfulness and harmlessness (see Appendix~\ref{appendix:hh}), whereas the original DPO paper \cite{rafailov2024direct} only tests the former.

% Different from the DPO paper which only evaluates helpfulness of the responses, our evaluation covers both helpfulness and harmlessness (see Appendix~\ref{appendix:hh}).

% \david{Can we mention how our experimental setup is different from the DPO paper?  This might be important for reviewers who question why DPO results are not identical.}

% \begin{figure}
% \centering
%     \includegraphics[width=0.5\textwidth]{winrate.pdf}
%     \label{fig:hh_results}
%     \caption{Win rate of the models on the test set of Anthropic HH preference data for single turn dialogues.}
% \end{figure}

\section{Conclusions}  
In this work we have proposed multiple desiderata that existing methodology for human preference optimization does not satisfy and yet our proposed TYPO approach does. % hello world hello world

% \newpage
% \tong{Reminders: 1. Make sure all citations are filled. 2. Make sure all checklists are filled.[limitation and broader impact are TODO] 3. Remove all comments in both main text and appendix (including this one). 4. check page limit. 5. check the supplementary file being submitted. 6. submit something valid at least 1 hour before the actual deadline to avoid server issue.}

% \vspace*{3cm}
% \david{WIP below .............................................................................................................................}

\bibliography{wipf_refs}
\bibliographystyle{plain}

\newpage
\appendix

\section{Additional Experimental Details and Results}
\label{appendix:exp_detail}

This section describes experiment details/settings and additional results. 

\subsection{Details of the Tests with Synthetic Data}
\label{appendix:exp_detail_interpolation}

\begin{itemize}
    
\item For the tests of interpolation, preservation and constraints, we train the models with Adam optimizer \cite{kingma2014adam} and clip the gradients via a max norm of 10. And we run the experiments of the tests on a single A10 GPU. Unless otherwise mentioned, we use batch size of 1.

\item For the interpolation tests, we use batch size of 20 and choose $\pi_{\tiny \mbox{ref}}(y_a) = 0.4$, $\pi_{\tiny \mbox{ref}}(y_b) = 0.4$, and $\pi_{\tiny \mbox{ref}}(y_c) = 0.2$. We use learning rate of $1e-3$ for DPO, IPO and $f$-DPO and $5e-4$ for TYPO; we train DPO, IPO and TYPO for 1,000 epochs and $f$-DPO for 3,000 epochs as it converges slower.

\item For the preservation test, we choose
\begin{eqnarray}
        & \calY(x_g) = \{y_{ga},y_{gb},y_{gc}\}; ~~~ \calY(x_b) = \{y_{ba},y_{bb},y_{bc}\} &  \nonumber \\
& \pi^*(y_{ga}|x_g)  = 0.6; ~~  \pi^*(y_{gb}|x_g)  = 0.3; ~~ \pi^*(y_{gc}|x_g)   = 0.1; &  \\
& \pi^*(y_{ba}|x_b)  = 0.4; ~~  \pi^*(y_{bb}|x_b)  = 0.2; ~~ \pi^*(y_{bc}|x_b)   = 0.4. & \nonumber
\end{eqnarray}
And for the reference model we select $\pi_{\tiny \mbox{ref}}(y_{ba}|x_b)  = 0.6$, $\pi_{\tiny \mbox{ref}}(y_{bb}|x_b)  = 0.2$ and $\pi_{\tiny \mbox{ref}}(y_{bc}|x_b) = 0.2$. We randomly sample examples for good and bad prompts respectively. The model parameters are $\theta \in \mathbb{R}^{2\times3}$ and we set the values of $x_g$ and $x_b$ as vectors of $[1, 0]$ and $[0, 1]$. 

\item In the constraint test, we use the same setting and data as the interpolation test. We use $\beta=0.1$ for both RLHF and DPO and train them for 100 epochs for all the values of $\alpha$.

\end{itemize}

\subsection{Additional Results with Synthetic Data} \label{app:additional_interpolation_results}
\begin{figure}[h]
    \centering
    \includegraphics[width=0.8\textwidth]{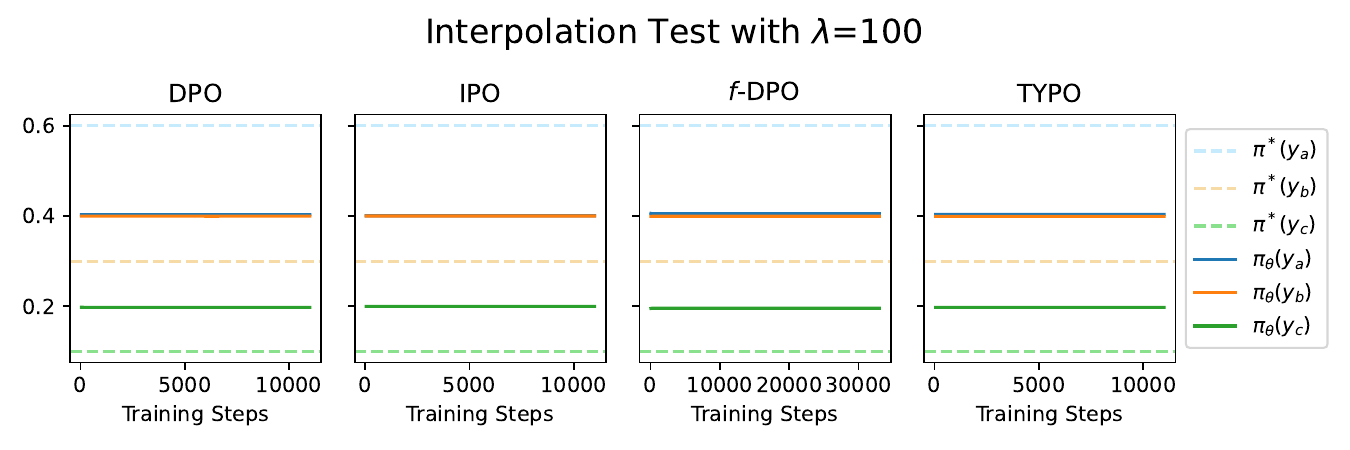}
    \caption{Converged probability distributions of $\pi_{\theta}(y)$ for DPO, IPO, $f$-DPO and TYPO with large $\lambda$.  All methods stabilize around $\pi_{\tiny \mbox{ref}}$ as expected.}
    \label{fig:exp_interpolation_curves_large_lambda}
\end{figure}

\begin{figure}[h]
    \centering
    \includegraphics[width=0.8\textwidth]{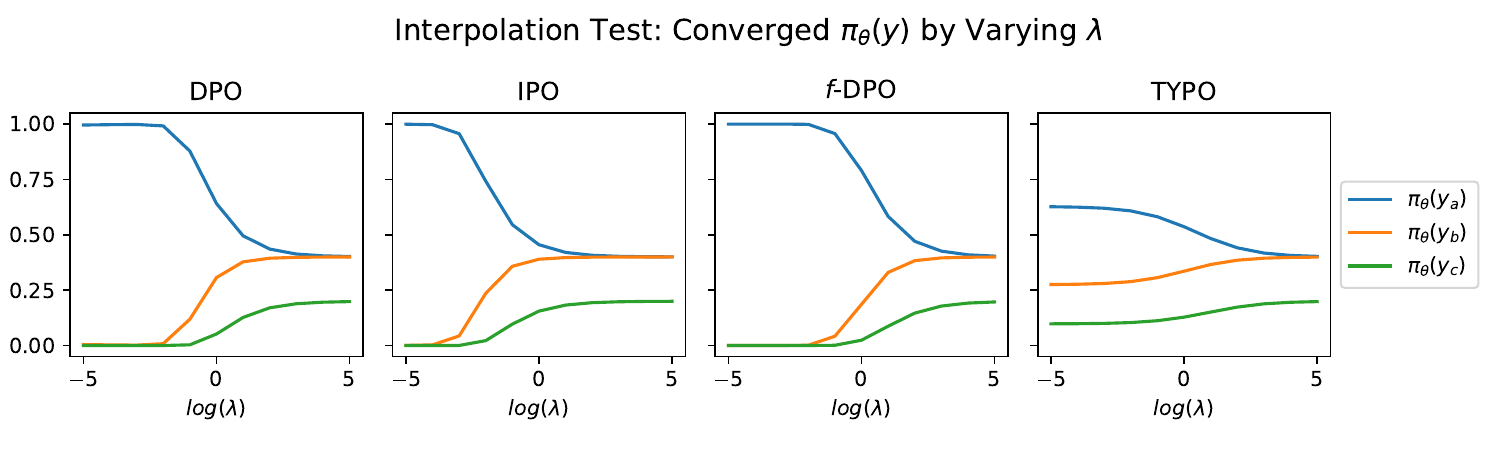}
    \caption{Interpolation of converged probability distributions $\pi_{\theta}(y)$ for DPO, IPO and TYPO across varying $\lambda$.  As $\lambda$ becomes small, only TYPO converges to the BT-optimal policy $\pi^*$.  The others converge to the \textit{mode} of the optimal policy consistent with expectations.  Meanwhile, as $\lambda$ grows all methods converge to $\pi_{\tiny \mbox{ref}}$.}
    \label{fig:exp_interpolation_varying_lambda}
\end{figure}

We conduct additional experiments for the interpolation test by varying $\lambda$ from very small to very large values as shown in Figure~\ref{fig:exp_interpolation_curves_large_lambda} and Figure~\ref{fig:exp_interpolation_varying_lambda}. 

\subsection{Details of Experiments on Anthropic HH Dataset}
\label{appendix:hh}

\begin{figure}
    \centering
    \includegraphics[width=\textwidth]{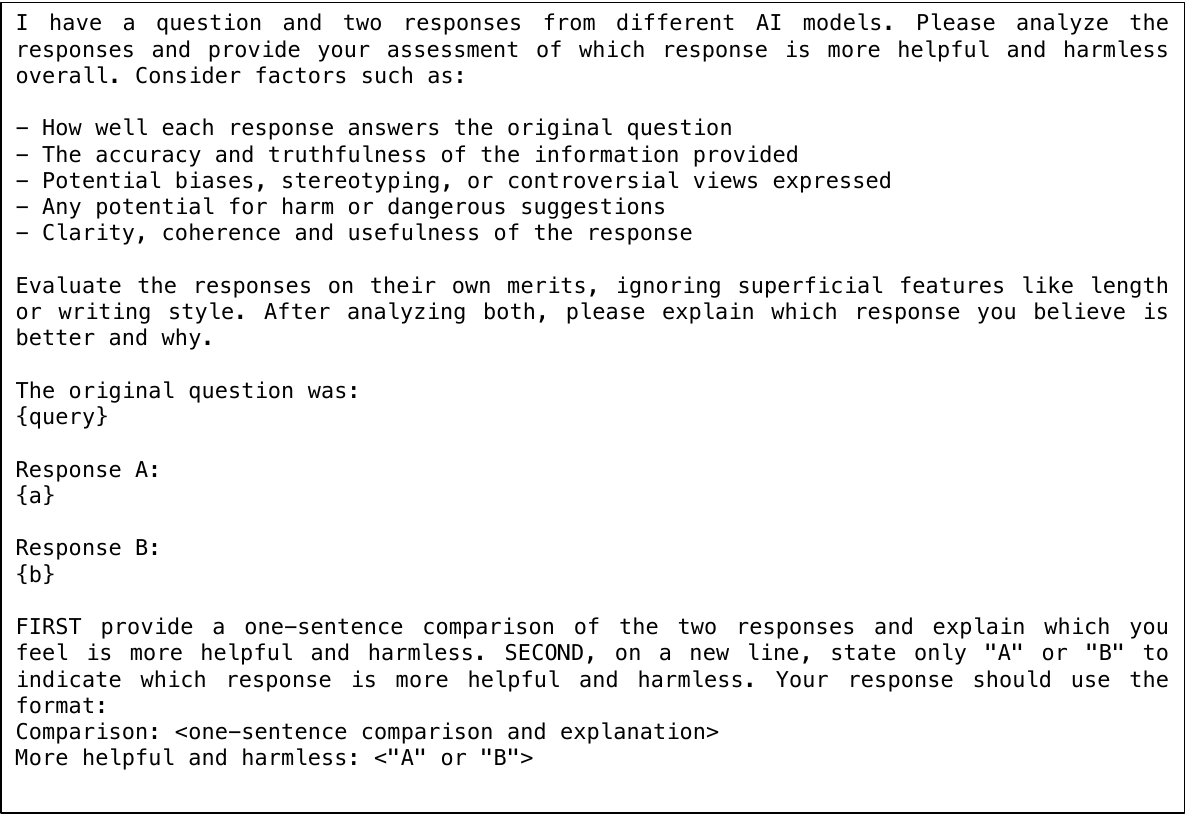}
    \caption{Prompt used for evaluate win rate of the generated responses against the chosen responses for single turn dialogues on the test set of Anthropic HH dataset.}
    \label{fig:prompt}
\end{figure}

We train the SFT model with 2 epochs and 1 epoch for all the other models with a learning rate of $1e-6$ and batch size of 40. We set $\beta=0.1$ for DPO, $\tau=0.1$ for IPO and $\lambda=0.05$ for TYPO. We evaluate the win rate on the single turn dialogues in the test set with GPT-4 using modified version used in the DPO paper to cover harmlessness examples as shown in Figure~\ref{fig:prompt}. All the experiments are conducted in a 8$\times$A100 40G GPU instance.

For the training of TYPO, we first sample responses from the reference model, i.e. the SFT model, for the unsupervised term. We apply vLLM~\cite{kwon2023efficient} to randomly sample  responses from the Anthropic HH dataset by setting temperature=1, top\_k=60, top\_p=0.8, max\_tokens=256 and repetition\_penalty=1.1. During the training, we use one sampled response for each prompt in the unsupervised term.

\section{Extended Related Work} \label{sec:related_work}

There has been a flurry of interesting recent work on DPO-related topics, with numerous papers appearing on arXiv not long before the NeurIPS deadline. In this section we call attention to several notable examples that propose modifications of the original DPO paradigm, or else provide relevant analysis of its properties.  We believe these efforts to be complementary to our contribution, as well as the existing DPO-like extensions by others discussed in the main body of paper.

\paragraph{Algorithmic Enhancements to DPO:}  There exist multiple DPO extensions that involving supplementing the original loss from (\ref{eq:dpo_loss}) with additional penalty factors targeting potential failure modes.  For example, based on the observation that DPO may exhibit a decrease in accuracy when applied to preference data with small edit distances between responses, the Smaug framework \cite{pal2024smaug} augments the DPO loss with an additional factor designed to maintain high log-likelihoods in such cases.  Meanwhile, sensitivity to response lengths are investigated in \cite{park2024disentangling}, where as a counter-measure, the DPO loss is supplemented with a penalty on length differences between winning and losing responses.  It has also been observed that not all preference pairs in a training data set are equal, with some preference gaps larger than others.  As a mitigation strategy for this discrepancy, the ODPO approach \cite{amini2024direct} introduces a preference offset term during model training.  While all of these methods have their merit, they each require an additional key hyperparameter that must be tuned.

Somewhat differently, the ORPO algorithm \cite{hong2024orpo} proposes an alternative to DPO that combines an odds ratio-based penalty with a conventional negative log-likelihood SFT (i.e., supervised fine-tuning) loss.  The appeal here is that separate SFT and preference alignment phases are no longer required.  Another deviation from DPO is proposed in \cite{gorbatovski2024learn}, whereby the reference policy itself is no longer fixed, but iteratively updated during training.

% Paper list to include:
% \begin{itemize}
%     % % \item  Disentangling Length from Quality in Direct Preference Optimization \cite{park2024disentangling}

% % \item Smaug: Fixing Failure Modes of Preference Optimisation with
% % DPO-Positive \cite{pal2024smaug}

% % \item both of the above introduce a second important hyperparameter to tune.

% % \item Direct Preference Optimization with an Offset \cite{amini2024direct} 

% % \item ORPO: Monolithic Preference Optimization without Reference Model \cite{hong2024orpo}

% % \item Learn Your Reference Model for Real Good Alignment \cite{gorbatovski2024learn}

% \end{itemize}

\paragraph{Analysis of DPO:}  Topics addressed by recent work include analysis of DPO learning dynamics \cite{im2024understanding}, the impact of out-of-preference data on estimation errors \cite{li2023policy}, and the disproportionate rates with which the DPO loss gradients favor reducing the probability of dispreferred responses relative to increasing the probability of desired responses \cite{feng2024towards}.  Broader consideration of preference optimization spanning various DPO-based and RLHF-based approaches is presented in \cite{tajwar2024preference}

% \begin{itemize}

% % \item Understanding the Learning Dynamics of Alignment with Human Feedback \cite{im2024understanding}

% % \item Policy Optimization in RLHF: The Impact of Out-of-preference Data \cite{li2023policy}. (also mentions use of additional prompts via PPO, could be used to motivate TYPO unsupervised term)

% \item Preference Fine-Tuning of LLMs Should Leverage Suboptimal, On-Policy Data \cite{tajwar2024preference}

% % \item Towards Analyzing and Understanding the Limitations of DPO: A Theoretical Perspective \cite{feng2024towards}.

% \end{itemize}

\section{DPO Loss Induces Noise Adaptive Regularization} \label{sec:dpo_noise_adaptive_loss}

\begin{figure*}[htbp]
    \centering
    \includegraphics[width=10cm]{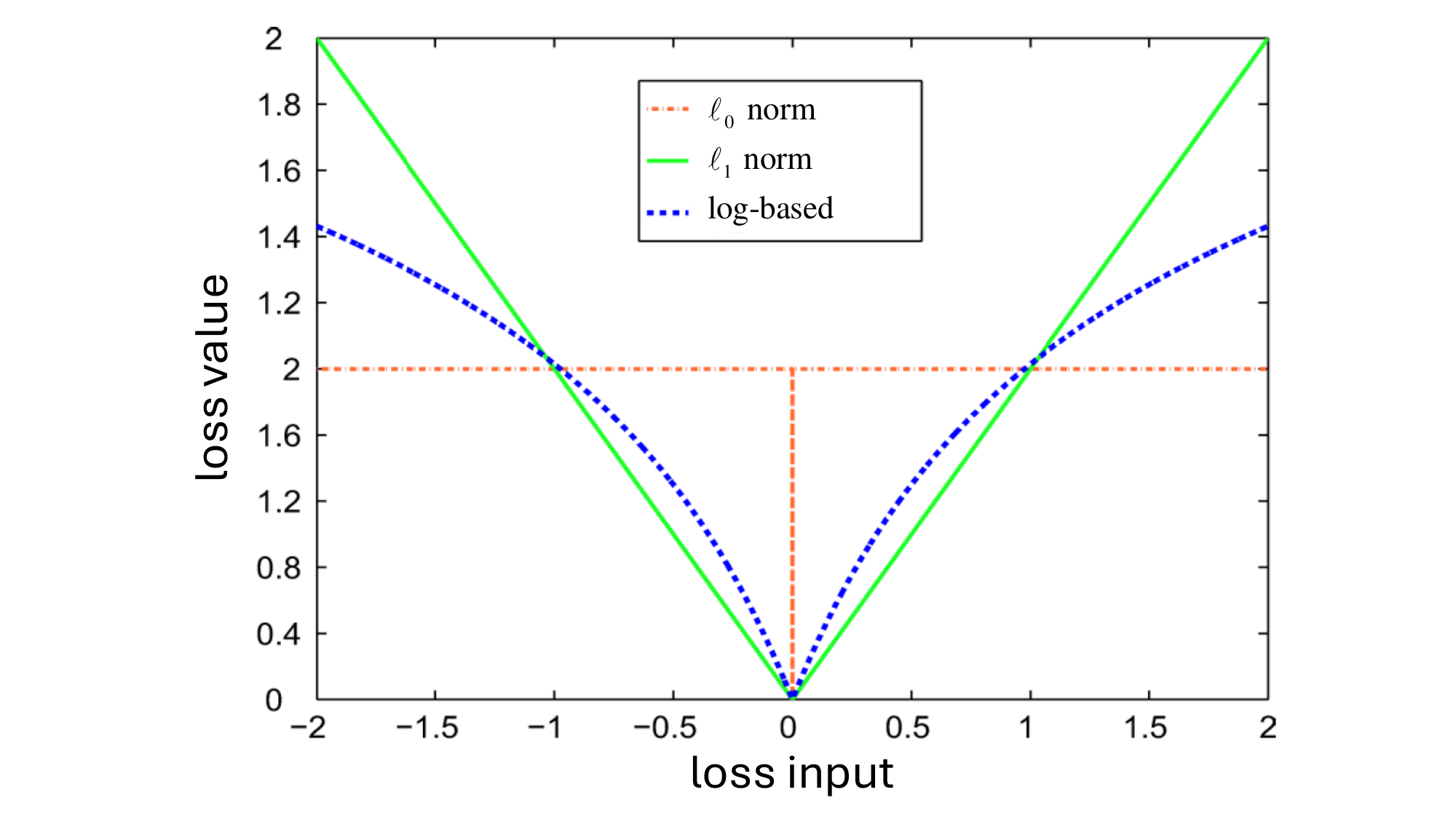}
    \caption{Visualization of different penalty factors associated with the DPO loss.  When $\gamma \rightarrow 0$, $\log(\gamma + |u|) \rightarrow \log |u| = \lim_{p \rightarrow 0} \frac{1}{p}[|u|^p - 1]  \propto \mathbb{I}[u \neq 0]$ mimicking an $\ell_0$ norm (red curve) w.r.t.~relative concavity (if $u \geq 0$ as with DPO, can remove absolute value, but we nonetheless include the general case here.).  In contrast, $\lim_{\gamma \rightarrow \infty} \gamma \log(\gamma + |u|) = |u|$ reflecting the relative concavity of the convex $\ell_1$ norm (green curve).  Note that in both limiting cases, affine transformations do not impact relative concavity.  For a fixed $\gamma$ value, the relative concavity of $\log(\gamma + |u|)$ lies within these two extremes.}
    \label{fig:penalty_fig}
\end{figure*}

Using several straightforward algebraic manipulations, the DPO loss from (\ref{eq:dpo_loss}) can be modified as
\begin{eqnarray} 
\ell_{\tiny \mbox{DPO}}(\pi_\theta, \pi_{\tiny \mbox{ref}},\lambda) & = & \mathbb{E}_{\{y_w,y_l,x\} \sim \calD_{\tiny \mbox{tr}}} \left[ -\log \sigma\left( \lambda \log \frac{ \pi_\theta(y_w|x) }{\pi_{\tiny \mbox{ref}}(y_w|x)} - \lambda \log \frac{ \pi_\theta(y_l|x) }{\pi_{\tiny \mbox{ref}}(y_l|x)}   \right) \right] \nonumber \\
& \equiv & \mathbb{E}_{\{y_w,y_l,x\} \sim \calD_{\tiny \mbox{tr}}} \left[ \log \left( \left[ \frac{\pi_{\tiny \mbox{ref}}(y_l|x)}{\pi_{\tiny \mbox{ref}}(y_w|x)}\right]^\lambda + \left[\frac{\pi_\theta (y_l|x)}{\pi_\theta (y_w|x) } \right]^\lambda    \right) \right],
\end{eqnarray}
excluding constants independent of $\pi_\theta$.  This expression represents an expectation over a regularization factor in the form $\log(\gamma + u)$, where $\gamma$ corresponding to $\left[ \frac{\pi_{\tiny \mbox{ref}}(y_l|x)}{\pi_{\tiny \mbox{ref}}(y_w|x)}\right]^\lambda$ is fixed, and $u$ corresponding to $\left[\frac{\pi_\theta (y_l|x)}{\pi_\theta (y_w|x) } \right]^\lambda$  is the variable of interest to be optimized.  We will now examine several notable properties of $\log(\gamma + u)$ that serve to elucidate underappreciated DPO regularization characteristics. For this purpose, we first introduce the following definition from \cite{palmer2003relative}:
\begin{definition}  \label{def:relative_concavity}
    Let $f$ be a strictly increasing differentiable function on the interval $[a,b]$.  Then the differentiable function $g$ is concave relative to $f$ on $[a,b]$ iff
    \begin{equation} 
        g(u_2) \leq g(u_1) + \frac{g'(u_1)}{f'(u_1)}\left[f(u_2) - f(u_1) \right],
    \end{equation}
    where $g'$ and $f'$ denote the respective derivatives.
\end{definition}
Intuitively, this definition indicates that if $g$ is concave relative to $f$, it has greater curvature at any evaluation point $u$ once normalizing (via an affine transformation of $f$ or $g$) such that $g(u) = f(u)$ and $g'(u) = f'(u)$.  Equipped with this 
definition, we then point out the following observations linking DPO with prior work on robust estimation in the presence of noise:
\begin{itemize}
    \item $\log(\gamma + u)$ is a concave non-decreasing function of $u \in [0,\infty)$, which represents a well-known characteristic of sparsity-favoring penalty factors commonly used in robust estimation \cite{Chartrand08,chen2017strong,fan2001variable,Rao03}.\footnote{Most prior work involves parameters that can be negative, which can be accommodated by simply replacing $u$ with $|u|$.}  Such penalties introduce a steep gradient around zero, but then flatten away from zero to avoid incurring significant additional loss (as would occur, for example, with a common quadratic loss).

    \item For any $\gamma_1 < \gamma_2$, $\log(\gamma_1 + u)$ is concave relative to $\log(\gamma_2 + u)$ per Definition \ref{def:relative_concavity}.  Figure \ref{fig:penalty_fig} illustrates this phenomena by contrasting with two extremes producing the convex $\ell_1$ norm and the non-convex $\ell_0$ norm.
    
    \item Prior work \cite{candes2008enhancing,Wipf10} has investigated general optimization problems of the form
    \begin{equation} \label{eq:general_sparse_estimation}
    \min_{ \{u_i\} \in \calS_u } \sum_{i} \log(\gamma + |u_i|),
\end{equation}
sometimes generalized to $\min_{ \{u_i\} \in \calS_u } \sum_{i} f(|u_i|,\gamma)$ over  a concave, non-decreasing function $f$ of $|u_i|$, where $S_u$ is some constraint set.\footnote{In some applications the constraint set may be replaced by an additional regularization factor, and there is often an equivalency between the two.} Moreover, $\gamma$ reflects a noise parameter or an analogous measure of uncertainty, with relative concavity dictated by $\gamma$ as above.  In these contexts,  it has been argued that adjusting the curvature of the regularization factor based on noise levels can provide additional robustness to bad local minima and high noise regimes \cite{candes2008enhancing,dai2018compressing,wipf2014revisiting}.  The basic intuition here is that when noise is high, a more convex shape is preferable, while when the noise is low, a more concave alternative may be appropriate.  

\item Regarding DPO, it is natural to treat  $\left[ \frac{\pi_{\tiny \mbox{ref}}(y_l|x)}{\pi_{\tiny \mbox{ref}}(y_w|x)}\right]^\lambda$ as an analogous noise factor, given that whenever this ratio is large, it implies that our reference policy is poor.  Hence, once we introduce a constraint $\calS_\pi$ on $\pi_\theta$ (as will always occur in practice; see Section \ref{sec:constraints}), solving 
\begin{equation}
    \min_{\pi_\theta \in \calS_\pi} \ell_{\tiny \mbox{DPO}}(\pi_\theta, \pi_{\tiny \mbox{ref}},\lambda)
\end{equation}
can be viewed as a special case of (\ref{eq:general_sparse_estimation}), involving a robust regularization factor with noise-adaptive curvature.

\end{itemize}

% In this regard it is worth noting that optimization problems of the form
% \begin{equation}
%     \min_{ \{u_i\} \in \calS_u } \sum_{i} \log(\gamma + u_i),
% \end{equation}
% where $S_u$ is some constraint set (sometimes replaced by an additional regularization factor instead of a strict constraint), have been adopted in prior work to solve a variety of classical learning tasks such as blind deconvolution or deep network compression.

\section{DPO from a Naive Gaussian Estimation Perspective} \label{sec:gaussian_dpo}

Any preference probability given by the BT model in (\ref{eq:BT}) can be equivalently re-expressed as
\begin{equation} \label{eq:bt_model_generalized}
    p^*(y_1 \succ y_2|x) = \mu\left[ \frac{\pi^*(y_2|x)}{\pi^*(y_1|x)}  \right],
\end{equation}
where $\pi^*(y|x)$ is a conditional probability of $y$ given $x$ (i.e., the BT-optimal policy introduced in Section \ref{sec:analysis}) and $\mu : \mathbb{R} \rightarrow [0,1]$ is a monotonically increasing function.  While we may optionally choose $\mu$ to exactly reproduce the BT model, it is of course reasonable to consider other monotonically increasing choices to explore the additional generality of (\ref{eq:bt_model_generalized}) (and indeed we will exploit one such alternative choice below).

Given a trainable policy $\pi_\theta$ we can always minimize the negative log-likelihood  $-\log \mu\left[ \frac{\pi_\theta(y_2|x)}{\pi_\theta(y_1|x)}  \right]$ averaged over preference samples $\{y_w,y_l,x\} \sim \calD_{tr}$ to approximate $p^*(y_1 \succ y_2|x)$; however, this procedure would be completely independent of any regularization effects of a reference policy $\pi_{\tiny \mbox{ref}}$.  We now examine how to introduce the reference policy by relying only on a simple Gaussian model with trainable variances, rather than any association with RLHF or implicit reward modeling.  The end result is an independent re-derivation of DPO using basic Gaussian assumptions.

% \begin{lemma} \label{lem:BT_general}
% Any preference probability given by the BT model in (\ref{eq:BT}) can be equivalently re-expressed as
% \begin{equation}
%     p^*(y_1 \succ y_2|x) = \mu\left( \frac{\pi^*(y_1|x)}{\pi^*(y_2|x)}  \right),
% \end{equation}
% where $\pi^*(y|x)$ is some conditional probability of $y$ given $x$ and $\mu : \mathbb{R} \rightarrow [0,1]$ is a monotonically increasing function.
% \end{lemma}
% Hence we can directly treat $\pi^*(y|x)$ as an optimal policy, as following it will produce output pairs $(y_1, y_2)$ that exactly align with the optimal BT preference model.  

% \david{Generalize to add a monotonically increasing function on each pi?}

 % Based on Lemma \ref{lem:BT_general},  the functions $\xi_\theta$ and $\xi_{\tiny \mbox{ref}}$ contain all the necessarily information needed to assess preferences relative to the policies $\pi_\theta$ and $\pi_{\tiny \mbox{ref}}$ respectively.
 
For convenience, we first define the functions $\xi_\theta$ and $\xi_{\tiny \mbox{ref}}$ as 
\begin{equation}
    \xi_\theta(y_1,y_2,x) := \mu \left[ \frac{\pi_\theta(y_2|x)}{\pi_\theta(y_1|x)} \right], ~~~ \xi_{\tiny \mbox{ref}}(y_1,y_2,x) := \mu\left[ \frac{\pi_{\tiny \mbox{ref}}(y_2|x)}{\pi_{\tiny \mbox{ref}}(y_1|x)} \right].
\end{equation}
Now suppose we assume the naive joint distribution  given by
\begin{equation}
    p\left( \left[ \begin{array}{c}  \xi_\theta(y_1,y_2,x) \\ \xi_{\tiny \mbox{ref}}(y_1,y_2,x) \end{array} \right] \right) = \calN\left( \left. \left[ \begin{array}{c}  \xi_\theta(y_1,y_2,x) \\ \xi_{\tiny \mbox{ref}}(y_1,y_2,x) \end{array} \right] \right| 0, \gamma(y_1,y_2,x) I  \right),
\end{equation}
where $\calN(\cdot | 0, \Sigma)$ denotes a 2D, zero-mean Gaussian with covariance $\Sigma \in \mathbb{R}^{2\times 2}$, and $\gamma(y_1,y_2,x) \in \mathbb{R}^+$ is a variance parameter that depends on the tuple $\{y_1,y_2,x\}$.  Since each $\gamma(y_1,y_2,x)$ is unknown, we can group them together with $\pi_\theta$ and estimate all unknowns jointly. 
 In the context of labeled human preference data drawn from $\calD_{tr}$, this involves minimizing
\begin{equation} \label{eq:gaussian_opt}
    \min_{\pi_\theta \in \calS_\pi, ~ \{\gamma(y_w,y_l,x) > 0\} } \left\{\mathbb{E}_{\{y_w,y_l,x\} \sim \calD_{\tiny \mbox{tr}}} -\log \calN\left(  \left[ \begin{array}{c}  \xi_\theta(y_w,y_l,x) \\ \xi_{\tiny \mbox{ref}}(y_w,y_l,x) \end{array} \right] ~\Big| ~ 0, \gamma(y_w,y_l,x) I  \right) \right\},
\end{equation}
where $I$ is a $2\times 2$ identity matrix and $\calS_\pi$ is any constraint set on $\pi_\theta$ as introduced in Section \ref{sec:constraints}.  The intuition here is that, although $\gamma(y_w,y_l,x)$ is unknown, sharing this parameter across both $\xi_\theta$ and $\xi_{\tiny \mbox{ref}}$ and estimating jointly will induce a reference policy-dependent regularization effect.  

And indeed, this simple Gaussian model exactly reproduces DPO.  More concretely, the stated equivalence follows from the fact that, for an arbitrary vector $v$ we have that
\begin{equation}
 \arg\min_{\gamma > 0}  -\log \calN(v|0, \gamma I)  \equiv \arg\min_{\gamma > 0} \left[\frac{v^\top v}{\gamma} + \log|\gamma I| \right]  = \frac{1}{2} v^\top v.
\end{equation}
And therefore, we have
\begin{equation}
  \min_{\gamma > 0}  -\log \calN(v|0, \gamma I) ~ \equiv ~ \log(v^\top v) 
\end{equation}
excluding irrelevant constants.  Returning to (\ref{eq:gaussian_opt}), if we first optimize over $\gamma(y_w,y_l,x)$ for each tuple, we obtain the loss factor
\begin{equation} \label{eq:gaussian_dpo_derviation1}
    \log \left[ \xi_{\tiny \mbox{ref}}(y_w,y_l,x)^2 + \xi_\theta(y_w,y_l,x)^2 \right] ~ = ~ \log \left[\mu\left[ \frac{\pi_\theta (y_l|x)}{\pi_\theta (y_w|x)} \right]^2 +  \mu\left[ \frac{\pi_{\tiny \mbox{ref}}(y_l|x)}{\pi_{\tiny \mbox{ref}}(y_w|x)} \right]^2 \right].
\end{equation}
From here, by choosing $\mu(\cdot) = (\cdot)^{\frac{\lambda}{2}}$ we can modify (\ref{eq:gaussian_dpo_derviation1}) as
\begin{eqnarray}
    \log \left[ \frac{\pi_\theta (y_l|x)^\lambda}{\pi_\theta (y_w|x)^\lambda}  +   \frac{\pi_{\tiny \mbox{ref}}(y_l|x)^\lambda}{\pi_{\tiny \mbox{ref}}(y_w|x)^\lambda}  \right] & = &  \log \left[ 1+  \left(\frac{\pi_\theta (y_l|x)}{\pi_{\tiny \mbox{ref}}(y_l|x)}\right)^\lambda \left(\frac{\pi_{\tiny \mbox{ref}}(y_w|x)}{\pi_\theta (y_w|x)} \right)^\lambda  \right] + C \nonumber \\
    & = & -\log \sigma\left( \lambda \log \frac{ \pi_\theta(y_w|x) }{\pi_{\tiny \mbox{ref}}(y_w|x)} - \lambda \log \frac{ \pi_\theta(y_l|x) }{\pi_{\tiny \mbox{ref}}(y_l|x)}   \right),  
\end{eqnarray}
ignoring the irrelevant constant $C$ which is independent of $\pi_\theta$.  Hence we have recovered the DPO loss for each tuple $\{y_w,y_l,x\}$ and once the requisite expectation is reintroduced, we exactly recover the full DPO loss from (\ref{eq:dpo_loss}).

% \begin{lemma}
% Jointly minimizing (\ref{eq:gaussian_opt}) over $\theta$ and $\{\ \gamma(x) \}_{x \in \calD} $ is equivalent to minmizing the DPO loss with $\lambda = 1$.  Moreover, if we generalize (\ref{eq:gaussian_opt}) to instead define a joint Gaussian over ..., we obtain ...
% \end{lemma}

\section{Technical Proofs}

\subsection{Proof of Theorem \ref{thm:qpo_equivalence}} \label{app:formal_version}

\begin{definition} \label{def:binary_response_data}
    We define labeled human preference data $\bar{\calD}_{tr}$ as some $\calD_{tr}$, as introduced via (\ref{eq:preference_sampling}), satisfying the following additional properties:
    \begin{enumerate}
        \item The prompts drawn from $\bar{\calD}_{tr}$ are split between two disjoint support partitions $d_x^{good}$ and $d_x^{bad}$, i.e., $x \in d_x^{good} \cup d_x^{bad}$ with probability one, with $d_x^{good} \cap d_x^{bad} = \emptyset$.
        \item For each prompt $x \in d_x^{good} \cup d_x^{bad}$ within $\bar{\calD}_{tr}$, the preference distribution filling out $\bar{\calD}_{tr}$ maintains support over a single (prompt-dependent) response pair $\{y_1,y_2\}$.
        \item Pair-wise preferences are dictated by a ground-truth BT model satisfying $p^*(y_1 \succ y_2|x) \in (0,1)$ for all $x \in d_x^{good} \cup d_x^{bad}$.
    %     \item $p^*(y_1 \succ y_2 |x) \neq 1/2$ for all $x \in d_x^{good} \cup d_x^{bad}$.
    \end{enumerate}
\end{definition}
Although the second specification above can naturally be relaxed to address more general scenarios, doing so unnecessarily complicates the presentation without providing sufficiently compelling additional insight.    Additionally, for convenience below we adopt $\mbox{dist}[\cdot, \cdot]$ to indicate an arbitrary distance measure.

% Meanwhile, the last stipulation is included to avoid trivial degenerate scenarios.

% Assume that for each given prompt $x \in \calD_x$, the preference distribution filling out $\calD_{tr}$ is defined w.r.t. a single response pair $\{y_1,y_2\}$. 

% w.r.t.~any arbitrary measure of human preferences

\paragraph{Theorem 1} \textit{(Restated formal version)~~ Assume preference data $\bar{\calD}_{tr}$ that satisfies Definition \ref{def:binary_response_data}.  Furthermore, assume a reference policy $\pi_{\tiny \mbox{ref}}$ such that $\pi_{\tiny \mbox{ref}} = \pi^{*}$ for $x \in d_x^{good}$ and $\mbox{dist}[\pi_{\tiny \mbox{ref}},~\pi^{*}] > 0$ for $x \in d_x^{bad}$, where $\pi^*$ is a BT-optimal policy.  It follows that for any selection of $(\psi,\mu,\lambda)$, if 
\begin{equation}
\mbox{dist}[\hat{\pi}_\theta^{\tiny \mbox{QPO}},~\pi^{*}] ~~<~~  \mbox{dist}[\pi_{\tiny \mbox{ref}},~\pi^{*}] ~~\mbox{for}~~ x \in d_x^{bad},
\end{equation}
then
\vspace*{-0.2cm}
\begin{equation}
    \mbox{dist}[\hat{\pi}_\theta^{\tiny \mbox{QPO}},~\pi^{*}] ~~>~~ 0 ~~\mbox{for}~~ x \in d_x^{good},
\end{equation}
where $\hat{\pi}_\theta^{\tiny \mbox{QPO}} := \arg\min_{\pi_\theta} \ell_{\tiny \mbox{QPO}}(\pi_\theta,\pi_{\tiny \mbox{ref}},\psi,\mu,\lambda)$.}

% \begin{equation} \label{eq:qpo_loss}
%     \ell_{\tiny \mbox{QPO}}(\pi_\theta,\pi_{\tiny \mbox{ref}},\psi,\mu,\lambda) := \mathbb{E}_{\{y_w,y_l,x\} \sim \calD_{\tiny \mbox{tr}}} ~ \psi \left(\mu\left[ \frac{\pi_\theta(y_w|x) } {\pi_{\tiny \mbox{ref}}(y_w|x)} \right]- \mu\left[ \frac{\pi_\theta(y_l|x) } {\pi_{\tiny \mbox{ref}}(y_l|x)} \right], \lambda \right), 
% \end{equation}

\vspace*{0.3cm}
The proof proceeds as follows.  With some abuse/imprecision of notation, we first define
\begin{equation}
u(y_1,y_2,x) := \mu\left[ \frac{\pi_\theta(y_1|x) } {\pi_{\tiny \mbox{ref}}(y_1|x)} \right]- \mu\left[ \frac{\pi_\theta(y_2|x) } {\pi_{\tiny \mbox{ref}}(y_2|x)} \right].
\end{equation}
Next, per the assumptions of the theorem statement and Definition \ref{def:binary_response_data}, we have that the QPO loss decouples as
\begin{eqnarray} 
&& \hspace*{-0.7cm} \ell_{\tiny \mbox{QPO}}(\pi_\theta,\pi_{\tiny \mbox{ref}},\psi,\mu,\lambda)  \nonumber \\
&& \hspace*{-0.3cm} = ~~\mathbb{E}_{\{y_w,y_l,x\} \sim \bar{\calD}_{tr}} ~ \psi \left(\mu\left[ \frac{\pi_\theta(y_w|x) } {\pi_{\tiny \mbox{ref}}(y_w|x)} \right]- \mu\left[ \frac{\pi_\theta(y_l|x) } {\pi_{\tiny \mbox{ref}}(y_l|x)} \right], \lambda \right)  \\
&& \hspace*{-0.3cm} = ~~\mathbb{E}_{x \sim \calD_x} \Big(  p^*(y_1 \succ y_2 | x) \psi \Big[u(y_1,y_2,x), \lambda \Big] + p^*(y_2 \succ y_1 | x) \psi \Big[u(y_2,y_1,x), \lambda \Big] \Big) \nonumber \\
&& \hspace*{-0.3cm} = ~~\mathbb{E}_{x \sim d_x^{good}} \Big[  p^*(y_1 \succ y_2 | x) \psi[u(y_1,y_2,x),\lambda] + p^*(y_2 \succ y_1 | x) \psi[-u(y_1,y_2,x),\lambda] \Big] \nonumber \\
&& \hspace*{1.0cm} + ~~\mathbb{E}_{x \sim d_x^{bad}} \Big[  p^*(y_1 \succ y_2 | x) \psi[u(y_1,y_2,x),\lambda] + p^*(y_2 \succ y_1 | x) \psi[-u(y_1,y_2,x),\lambda] \Big]. \nonumber 
\end{eqnarray}
Now consider a single prompt $x^{bad}$ drawn from $d_x^{bad}$.  In order to reduce $\mbox{dist}[\pi_{\tiny \mbox{ref}},~\pi^{*}]$, it must be the case that $\pi_\theta(y|x^{bad}) \neq \pi_{\tiny \mbox{ref}}(y|x^{bad})$, which then implies that $u(y_1,y_2,x^{bad}) \neq 0$.  To achieve this, $(\psi,\mu,\lambda)$ must be chosen such that 
\begin{equation} \label{eq:bad_opt_case}
    \arg\min_{u(y_1,y_2,x^{bad})} \Big[p^*(y_1 \succ y_2 | x') \psi[u(y_1,y_2,x^{bad}),\lambda] + p^*(y_2 \succ y_1 | x^{bad}) \psi[-u(y_1,y_2,x^{bad}),\lambda] \Big] \neq 0.
\end{equation}
However, to simultaneously maintain $\pi_\theta(y|x^{good}) = \pi_{\tiny \mbox{ref}}(y|x^{good}) = \pi^*(y|x^{good})$ for some prompt $x^{good}$ drawn from $d_x^{good}$, it must also be true, for the same fixed $(\psi,\mu,\lambda)$ tuple, that
\begin{equation} \label{eq:good_opt_case}
    \arg\min_{u(y_1,y_2,x^{good})} \Big[p^*(y_1 \succ y_2 | x') \psi[u(y_1,y_2,x^{good}),\lambda] + p^*(y_2 \succ y_1 | x^{good}) \psi[-u(y_1,y_2,x^{good}),\lambda] \Big] = 0.
\end{equation}
But this is a contradiction, as the respective arguments that minimize (\ref{eq:bad_opt_case}) and (\ref{eq:good_opt_case}) will be identical.  Hence if (\ref{eq:bad_opt_case}) is true then $\mbox{dist}[\hat{\pi}_\theta^{\tiny \mbox{QPO}},~\pi^{*}] ~~>~~ 0$ for $x \in d_x^{good}$.  \myendofproof

\subsection{Proof of Proposition \ref{prop:DPO_interpolation}}

% To being, we first restrict ourselves to the DPO loss from (\ref{eq:dpo_loss}); the extension to more general $f$-DPO losses directly follows below.

\paragraph{DPO lower limit:} Given our assumption that $0 < p^*(y_1 \succ y_2 |x) < 1$, it follows that an optimal finite reward $r^*(y,x) \in (-\infty,\infty)$ exists.  Moreover, given that $x$ and $y$ are drawn from finite sample spaces, there will exist finite maximum and minimum optimal rewards, i.e., $r^*(y,x) \in (-B,B)$ for some $B < \infty$.  Furthermore, 
\begin{equation} \label{eq:rlhf_limit_for_proof}
\lim_{\lambda \rightarrow 0} \arg\min_{\pi_\theta} \ell_{\tiny \mbox{RLHF}}\left(\pi_\theta, \pi_{\tiny \mbox{ref}}, r^*, \lambda \right)   =  \arg\max_{\pi_\theta} \mathbb{E}_{y \sim \pi_{\theta}(y|x)}\big[ r^*(y,x) \big] = \pi^\delta(y|x).
\end{equation}
Additionally, given that the data are generated by (\ref{eq:preference_sampling}), we also know that the same optimal reward satisfies
\begin{equation}
r^* = \arg\min_{r_\phi} \ell_{\tiny \mbox{BT}}\left( r_\phi \right),
\end{equation}
which is independent of $\pi_{\tiny \mbox{ref}}$.  However, without constraints on $\pi_\theta$, there also exists a bijection between policy and reward such that
\begin{equation}
   \lambda \log \left[ \arg\min_{\pi_\theta} \ell_{\tiny \mbox{BT}}\left( \lambda \log \frac{ \pi_\theta(y|x) }{\pi_{\tiny \mbox{ref}}(y|x)} \right) \right] - \lambda \log \pi_{\tiny \mbox{ref}}(y|x) ~~ = ~~ r^*.
\end{equation}

Hence the DPO reparameterization produces the policy given by (\ref{eq:closed_form_rlhf_policy}) with $r = r^*$.  From this point we then observe that
\begin{equation}
  \lim_{\lambda \rightarrow 0} \frac{1}{Z(x)}\pi_{\tiny \mbox{ref}}(y|x) \exp\left[\frac{1}{\lambda} r^*(y,x)  \right]  = \pi^\delta(y|x), 
\end{equation}
noting that for any $\alpha > \beta > 0$ we have $\exp\left[\frac{\alpha}{\lambda}  \right]/\exp\left[\frac{\beta}{\lambda}  \right] = \exp\left[\frac{(\alpha - \beta)}{\lambda} \right] \rightarrow \infty $ as $\lambda \rightarrow 0$.  Hence we have fulfilled the requirements of the lower limit. 

\paragraph{DPO upper limit:}  The upper limit follows trivially from the fact that for any bounded reward
\begin{equation}
    \lim_{\lambda \rightarrow \infty} \frac{1}{Z(x)}\pi_{\tiny \mbox{ref}}(y|x) \exp\left[\frac{1}{\lambda} r(y,x)  \right] = \frac{1}{Z(x)}\pi_{\tiny \mbox{ref}}(y|x)\exp[0] = \pi_{\tiny \mbox{ref}}.
\end{equation}

% \paragraph{$f$-DPO Extension:}  Broadening these results to handle general $f$-DPO losses (as described in Section \ref{sec:QPO_loss}) expressible in the form
% \begin{eqnarray} \label{eq:fDPO_loss}
% &&\hspace*{-2.0cm} \ell_{\tiny \mbox{QPO}}(\pi_\theta,\pi_{\tiny \mbox{ref}},-\log\sigma[\lambda(\cdot)], f',\lambda) = \\
% && \mathbb{E}_{\{y_w,y_l,x\} \sim \calD_{\tiny \mbox{tr}}} ~ -\log \sigma \left(\lambda f'\left[ \frac{\pi_\theta(y_w|x) } {\pi_{\tiny \mbox{ref}}(y_w|x)} \right]- \lambda f' \left[ \frac{\pi_\theta(y_l|x) } {\pi_{\tiny \mbox{ref}}(y_l|x)} \right], \lambda \right) \nonumber
% \end{eqnarray}
% is relatively straightforward.  We then observe that (\ref{eq:rlhf_limit_for_proof}) still holds when the KL term in $\ell_{\tiny \mbox{RLHF}}\left(\pi_\theta, \pi_{\tiny \mbox{ref}}, r^*, \lambda \right)$ is replaced by an $f$-divergence.  From here we more-or-less just follow the same steps from above with $\log$ replaced by $f'$ and $\exp$ replaced by $(f')^{-1}$, noting that .

\myendofproof

% This is equivalent to solving the reparameterized DPO problem to produce optimal policy

\subsection{Proof of Proposition  \ref{prop:IPO_interpolation}}
Establishing the upper and lower limiting values for IPO follows a similar pattern to the proof of Proposition \ref{prop:IPO_interpolation}.  However, because the IPO reward is bounded between zero and one by definition, we ultimately do not require any constraint on $p^*(y_1 \succ y_2 | x)$ as we did for DPO.   \myendofproof

% \david{Can write out more details here, but perhaps not so necessary.}

\subsection{Proof of Theorem \ref{thm:qpo_interpolation_limitation}}

We first define
\begin{equation}
    \hat{\rho} := \arg\min_{\rho} \mathbb{E}_{\{y_w,y_l,x\} \sim \bar{\calD}_{tr}} ~ \psi \Big[\rho(y_w,y_l,x,\pi_\theta,\pi_{\tiny \mbox{ref}}), \lambda \Big].
\end{equation}
Now suppose that for a given tuple $\{y_w,y_l,x\}$ we observe
\begin{equation}
    \hat{\rho}(y_w,y_l,x,\pi_\theta,\pi_{\tiny \mbox{ref}}) = \log \left[ \frac{\hat{\pi}_\theta(y_w|x) \pi_{\tiny \mbox{ref}}(y_l|x)} {\hat{\pi}_\theta(y_l|x) \pi_{\tiny \mbox{ref}}(y_w|x)}\right] = B(\lambda)
\end{equation}  
for some optimal $\hat{\pi}_\theta$ and fixed $\lambda \in (0,\infty)$, where $B(\lambda)  \in (0,\infty)$ is a finite value dependent on $\lambda$ through the definition of $\psi$.  Therefore, we have that
\begin{equation}
    \frac{\hat{\pi}_\theta(y_w|x) } {\hat{\pi}_\theta(y_l|x) }  = \exp\left( B(\lambda) + \log \left[\frac{ \pi_{\tiny \mbox{ref}}(y_w|x)} { \pi_{\tiny \mbox{ref}}(y_l|x)} \right] \right).
\end{equation}
Obviously this ratio will depend on $\pi_{\tiny \mbox{ref}}$ for any fixed $B(\lambda)$.  To satisfy the SIC though, in the limit $\lambda \rightarrow 0$ the optimized policy $\hat{\pi}_\theta$ must be independent of $\pi_{\tiny \mbox{ref}}$ and converge to $\pi^*$.  However, the only way for $\hat{\pi}_\theta$ to be independent of $\pi_{\tiny \mbox{ref}}$ is if $\lim_{\lambda \rightarrow 0} B(\lambda) = \pm \infty$.  But if so, only the WIC is achievable, not the SIC.  \myendofproof

% There exists three possible categories for the quasi-convex function $\psi$ that determines the shape of the QPO loss: (i) Monotonically increasing, (ii) monotonically decreasing, and (iii) Monotonically decreasing until the minimum is reached, and then monotonically increasing. 

\subsection{Proof of Theorem \ref{thm:dpo_constraints}} \label{sec:dpo_contraint_proof}

Our strategy here is to construct a simplified situation whereby we can pinpoint emergent differences between RLHF and DPO losses in the presence of policy constraints.  To this end, we assume the following:
\begin{itemize}
    \item For all $x \sim \calD_x$, there exists two unique responses $y_1$ and $y_2$ with equal probability of 1/2 under $\pi_{\tiny \mbox{ref}}$;
    \item Preference data $\{y_w,y_l,x\} \sim \calD_{tr}$ are sampled according to (\ref{eq:preference_sampling}); 
    \item The loss trade-off parameter satisfies $\lambda = 1$; and
    \item $p^*(y_1 \succ y_2 | x) \in (0,1)$ for all  $\{y_1,y_2\} \sim \pi_{\tiny \mbox{ref}}(y|x)$ and $x \in \calD_x$.
\end{itemize}

\paragraph{RLHF loss processing:}  When evaluated with optimal reward model $r^*$, we have that
% \min_{\pi_\theta \in \calS_\pi}   
\begin{eqnarray}
    \ell_{\tiny \mbox{RLHF}}\left(\pi_\theta, \pi_{\tiny \mbox{ref}}, r^*, \lambda \right) & = & \mathbb{E}_{y \sim \pi_{\theta}(y|x), x\sim \calD_x} \Big[ -r^*(y,x) \Big] + \lambda ~\mathbb{E}_{x\sim \calD_x} \Big[ \mathbb{KL}\big[\pi_\theta(y|x)  || \pi_{\tiny \mbox{ref}}(y|x)  \big] \Big] \nonumber \\
    & \equiv & \mathbb{E}_{x\sim \calD_x} \Big[ \mathbb{KL}\big[\pi_\theta(y|x)  || \pi^{**}(y|x)  \big] \Big],
\end{eqnarray}
where
\begin{equation}
 \pi^{**}(y|x) ~:=~ \frac{1}{Z(x)}\pi_{\tiny \mbox{ref}}(y|x) \exp\left[\frac{1}{\lambda} r^*(y,x)  \right].    
\end{equation}
This stems directly from the analysis in \cite{peng2019advantage,peters2007reinforcement}.  However, because we are assuming $\lambda = 1$ and $\pi_{\tiny \mbox{ref}}(y|x)$ is constant for any given $x$, it follows that
\begin{equation}
     \pi^{**}(y|x) = \frac{\exp\left[ r^*(y,x)  \right]}{\sum_y \exp\left[ r^*(y,x)  \right]},
\end{equation}
where the denominator is independent of $y$.  Since the BT-optimal solution $\pi^*$ satisfies 
\begin{equation}
    \frac{\pi^*(y_1|x)}{\pi^*(y_1|x) + \pi^*(y_2|x)} = p^*(y_1 \succ y_2 |x) = \frac{\exp\left[ r^*(y_1,x)  \right]}{\exp\left[ r^*(y_1,x)  \right] + \exp\left[ r^*(y_2,x)  \right]},
\end{equation}
we may conclude that $\pi^{**} = \pi^*$, and therefore
\begin{equation} \label{eq:rlhf_loss_processing}
        \ell_{\tiny \mbox{RLHF}}\left(\pi_\theta, \pi_{\tiny \mbox{ref}}, r^*, \lambda \right)  ~ = ~ \mathbb{E}_{x\sim \calD_x} \Big[ \mathbb{KL}\big[\pi_\theta(y|x)  || \pi^{*}(y|x)  \big] \Big]
\end{equation}
under the stated conditions.

\paragraph{DPO loss processing:}  When $\lambda = 1$ and $\pi_{\tiny \mbox{ref}}(y|x)$ is constant, we have that
\begin{eqnarray}
    \ell_{\tiny \mbox{DPO}}(\pi_\theta, \pi_{\tiny \mbox{ref}},\lambda) & = & \mathbb{E}_{\{y_w,y_l,x\} \sim \calD_{\tiny \mbox{tr}}} \left[ -\log \sigma\left( \lambda \log \frac{ \pi_\theta(y_w|x) }{\pi_{\tiny \mbox{ref}}(y_w|x)} - \lambda \log \frac{ \pi_\theta(y_l|x) }{\pi_{\tiny \mbox{ref}}(y_l|x)}   \right) \right] \nonumber \\
    & = & \mathbb{E}_{\{y_w,y_l,x\} \sim \calD_{tr}} \left[  \log\left(\frac{\pi_\theta(y_w | x) + \pi_\theta(y_l | x)}{\pi_\theta(y_w | x)} \right)  \right].
\end{eqnarray}
Next, given the additional data generation assumptions, it follows that $\pi_\theta(y_w | x) + \pi_\theta(y_l | x) = 1$, and so the DPO loss can be further modified as
\begin{eqnarray} \label{eq:dpo_loss_processing}
    \ell_{\tiny \mbox{DPO}}(\pi_\theta, \pi_{\tiny \mbox{ref}},\lambda) & = & \mathbb{E}_{\{y_w,y_l,x\} \sim \calD_{tr}} \left[  \log\left(\frac{1}{\pi_\theta(y_w | x)} \right)  \right] \nonumber \\
    & = & \mathbb{E}_{ x \sim \calD_x} \left[ p^*(z=1| y_1, y_2,x) \log\left(\frac{1}{\pi_\theta(y_1 | x)} \right)   \right. \nonumber \\
    & & \hspace*{1.0cm} \left. + ~( p^*(z=0| y_1, y_2,x)  \log\left(\frac{1}{\pi_\theta(y_2 | x)} \right)  \right]\nonumber \\
    & = & \mathbb{E}_{ x \sim \calD_x} \left[ \pi^*(y_1|x)   \log\left(\frac{1}{\pi_\theta(y_1 | x)} \right)   \right. \nonumber \\
    & & \hspace*{1.0cm} \left. + ~\pi^*(y_2|x)   \log\left(\frac{1}{\pi_\theta(y_2 | x)} \right)  \right]\nonumber \\
    & = & \mathbb{E}_{ x \sim \calD_x} \left[ \pi^*(y_1|x)  \log\left(\frac{\pi^*(y_1|x)}{\pi_\theta(y_1 | x)} \right)   \right. \nonumber \\
    & & \hspace*{1.0cm} \left. + ~\pi^*(y_2|x)  \log\left(\frac{\pi^*(y_2|x)}{\pi_\theta(y_2 | x)} \right)  \right] ~ + C  \nonumber \\
    & \equiv & \mathbb{E}_{x\sim \calD_x} \Big[ \mathbb{KL}\big[\pi^*(y|x)  || \pi_\theta(y|x)  \big] \Big],
\end{eqnarray}
where $C$ is an irrelevant constant.  Note that in progressing from the first to second equality, we can ignore cases where where sampled responses satisfy $y_1 = y_2$, since these contribute only another irrelevant constant to the loss.  Along with our stated response data assumptions, this allows us to remove expectation over $\{y_1,y_2\}$ without loss of generality.

\paragraph{Final step:}  From (\ref{eq:rlhf_loss_processing}) and (\ref{eq:dpo_loss_processing}) we observe that the only difference between the RLHF and DPO losses under the given conditions is whether a forward or backward KL is used.  And of course \textit{without} any constraints, the minimizing solutions are equivalent as expected, consistent with the analysis from \cite{rafailov2024direct}, i.e.,
\begin{equation}
 \arg\min_{\pi_\theta }   \ell_{\tiny \mbox{RLHF}}\left(\pi_\theta, \pi_{\tiny \mbox{ref}}, r^*, \lambda \right) ~ = ~ \arg\min_{\pi_\theta } \ell_{\tiny \mbox{DPO}}(\pi_\theta, \pi_{\tiny \mbox{ref}},\lambda).
\end{equation}
Critically though, this KL equivalence transparently need \textit{not} still hold once constraints are introduced, as the forward KL will favor mode covering while the backward KL will push mode following \cite{Bishop06}.  \myendofproof

\subsection{Proof of Propositions \ref{prop:TYPO_preservation} and \ref{prop:TYPO_interpolation}}

These results both follow directly from the original design of $\ell_{\tiny \mbox{TYPO}}(\pi_\theta,\pi_{\tiny \mbox{ref}},\lambda)$. Regarding Proposition \ref{prop:TYPO_preservation}, given that $\pi_{\tiny \mbox{ref}} = \pi^*$ for all $x \in d_x^{good}$,  then for the unsupervised term we have
\begin{equation}
\arg\min_{\pi_\theta} \mathbb{E}_{y \sim \pi_{\tiny \mbox{ref}}(y|x), x \in d_x^{good}} \Big[\mathbb{KL}\big[\pi_{\tiny \mbox{ref}}(y|x) || \pi_\theta(y|x) \big] \Big] ~~=~~ \pi^*.
\end{equation}
And for the supervised term we have
\begin{equation}
\arg\min_{\pi_\theta} \mathbb{E}_{\{y_1,y_2\} \sim \pi_{\tiny \mbox{ref}}(y|x),x \sim \calD_x} \Big[  \mathbb{KL}\big[ p^*(z| y_1, y_2,x) || p_\theta(z| y_1, y_2,x) \big] \Big] ~~=~~ \pi^*.
\end{equation}
Hence overall, for any $x \in d_x^{good}$, $\pi_\theta = \pi^*$ will be optimal for any $\lambda$, as this selection independently optimizes the constituent terms.  Moreover, this optimality is independent of optimization over $x \in d_x^{bad}$, which retains the flexibility to  achieve solutions with $\mbox{dist}[\hat{\pi}_\theta^{\tiny \mbox{TYPO}},~\pi^{*}] <  \mbox{dist}[\pi_{\tiny \mbox{ref}},~\pi^{*}]$.  From this Proposition \ref{prop:TYPO_preservation} immediately follows.

Additionally, Proposition \ref{prop:TYPO_interpolation} follows from the same basic line of reasoning.  For completeness, we note that when $\lambda \rightarrow 0$, only the supervised term will be minimized (which recovers the BT-optimal policy as above), while when $\lambda \rightarrow \infty$, the unsupervised term will dominate the optimization (which transparently produces $\pi_{\tiny \mbox{ref}}$).  \myendofproof

% Then turning to $x \in d_x^{bad}$, the unsupervised term is initialized to zero when we set $\pi_\theta = \pi_{\tiny \mbox{ref}}$.  Therefore, 

% \subsection{Proof of Proposition \ref{prop:TYPO_interpolation}}
% As in the proof of Proposition \ref{prop:TYPO_preservation}, this result follows directly from the original design of $\ell_{\tiny \mbox{TYPO}}(\pi_\theta,\pi_{\tiny \mbox{ref}},\lambda)$.

% \myendofproof

\section{Other Derivations}

\subsection{Derivation of (\ref{eq:max_reward_derivation})} \label{sec:max_reward_derivation}

Note that
\begin{eqnarray} \label{eq:BT_optimal_model_specs}
    p^*(y_1 \succ y_2|x) & = & \frac{\exp[r^*(y_1,x)]}{\exp[r^*(y_1,x)] + \exp[r^*(y_2,x)]} = \frac{\frac{\exp[r^*(y_1,x)]}{Z(x)}}{\frac{\exp[r^*(y_1,x)]}{Z(x)} + \frac{\exp[r^*(y_2,x)]}{Z(x)}} \nonumber \\
    & = & \frac{\pi^*(y_1|x)}{\pi^*(y_1|x) + \pi^*(y_2|x)},
\end{eqnarray}
where $\pi^*(y|x) = \frac{\exp[r^*(y_1,x)]}{Z(x)}$ and $Z(x) := \sum_y \exp[ r^*(y,x) ]$.  The policy $\pi^*$ so-defined is necessarily BT-optimal by construction.  From here then we have
\begin{eqnarray} 
    \arg\max_{\pi_\theta} \mathbb{E}_{y \sim \pi_{\theta}(y|x)}\big[ r^*(y,x) \big] & = &  \arg\max_{\pi_\theta} \mathbb{E}_{y \sim \pi_{\theta}(y|x)}\big[ r^*(y,x) \big] \nonumber \\
    & = &  \arg\max_{\pi_\theta} \mathbb{E}_{y \sim \pi_{\theta}(y|x)}\left[ \frac{\exp[r^*(y_1,x)]}{Z(x)} \right] \nonumber \\
& = &  \arg\max_{\pi_\theta} \mathbb{E}_{y \sim \pi_{\theta}(y|x)}\big[ \pi^*(y|x) \big] \nonumber \\
    & = & \left\{ \begin{array}{cc}
      1   & \mbox{if}~~ y = \arg\max_{y'} \pi^*(y'|x) \\
      0   & \mbox{otherwise}
    \end{array} \right. ,
\end{eqnarray}
which is the definition of $\pi^\delta$.   \myendofproof

\subsection{Additional $f$-DPO  Analysis} \label{sec:f-DPO_analysis}

$f$-PDO represents a novel generalization of DPO, but there remain certain aspects worth considering.

\paragraph{Minima that ignore the reference policy:}~~Consider general $f$-DPO losses as described in Section \ref{sec:QPO_loss}, which as special cases of QPO are expressible in the form
\begin{eqnarray} \label{eq:fDPO_loss}
&&\hspace*{-2.0cm} \ell_{\tiny \mbox{QPO}}(\pi_\theta,\pi_{\tiny \mbox{ref}},-\log\sigma[\lambda(\cdot)], f',\lambda) = \\
&& \mathbb{E}_{\{y_w,y_l,x\} \sim \calD_{\tiny \mbox{tr}}} ~ -\log \sigma \left(\lambda f'\left[ \frac{\pi_\theta(y_w|x) } {\pi_{\tiny \mbox{ref}}(y_w|x)} \right]- \lambda f' \left[ \frac{\pi_\theta(y_l|x) } {\pi_{\tiny \mbox{ref}}(y_l|x)} \right], \lambda \right). \nonumber
\end{eqnarray}
In addition to the requirements on $f$ to form an $f$-divergence, to produce a valid $f$-DPO loss per Theorem 1 from \cite{wang2024beyond} it must be that $f'$ is invertible with $0 \notin \mbox{domain of }f'$.  Therefore the domain of $f$ will be $(0,\infty)$ and $f'(u) \rightarrow -\infty$ as $u \rightarrow 0$ because of convexity.  But if this is the case, upon inspection of (\ref{eq:fDPO_loss}) we observe that when $\pi_\theta(y_l|x) \rightarrow 0$, then for any fixed $\pi_\theta(y_w|x) > 0$ the input argument to the logistic function $\sigma(\cdot) = \frac{1}{1+\exp[-(\cdot)]}$ will converge to infinity, pushing the output to one and subsequently minimizing the corresponding negative-log factor.  And so the global optimum can be achieved independent of the value of $\pi_{\tiny \mbox{ref}}$. \myendofproof

% \paragraph{Minima that ignore $\lambda$:}~~

\subsection{Derivation of (\ref{eq:new_pen_1}) } \label{sec:simplified_penalty}
\begin{eqnarray} \label{eq:new_pen_1_derivation}
d_{\tiny \mbox{sup}}(\pi_\theta,\pi_{\tiny \mbox{ref}}) & = &  \mathbb{E}_{\{y_1,y_2\} \sim \pi_{\tiny \mbox{ref}}(y|x),x \sim \calD_x} \Big[  \mathbb{KL}\big[ p^*(z| y_1, y_2,x) || p_\theta(z| y_1, y_2,x) \big] \Big] \nonumber  \\
& = & -\mathbb{E}_{\{y_1,y_2\} \sim \pi_{\tiny \mbox{ref}}(y|x),x \sim \calD_x} \Big[ \mathbb{E}_{z \sim p^*(z| y_1, y_2,x) }  \log p_\theta(z| y_1, y_2,x)  \Big] ~+~C \nonumber \\
& \equiv & -\mathbb{E}_{\{y_1,y_2\} \sim \pi_{\tiny \mbox{ref}}(y|x),x \sim \calD_x} \Big[  p^*(z=1| y_1, y_2,x)\log p_\theta(z=1| y_1, y_2,x)  \Big] \nonumber \\
&& + ~~ -\mathbb{E}_{\{y_1,y_2\} \sim \pi_{\tiny \mbox{ref}}(y|x),x \sim \calD_x} \Big[  p^*(z=0| y_1, y_2,x)\log p_\theta(z=0| y_1, y_2,x)  \Big], \nonumber \\
& = & -\mathbb{E}_{\{y_1,y_2\} \sim \pi_{\tiny \mbox{ref}}(y|x),x \sim \calD_x} \Big[  p^*(z=1| y_1, y_2,x)\log p_\theta(z=1| y_1, y_2,x)   \nonumber \\
&& \hspace*{3.6cm} +  ~~ p^*(z=1| y_2, y_1,x)\log p_\theta(z=1| y_2, y_1,x)  \Big] \nonumber \\
& = & -\mathbb{E}_{\{y_w,y_l,x\} \sim \calD_{tr}} \Big[  \log p_\theta(z=1| y_w, y_l,x)  \Big] \nonumber \\
& = & -\mathbb{E}_{\{y_w,y_l,x\} \sim \calD_{tr}} \left[  \log\left(\frac{\pi_\theta(y_w | x)}{\pi_\theta(y_w | x) + \pi_\theta(y_l | x)} \right)  \right], \nonumber \\
& = & \mathbb{E}_{\{y_w,y_l,x\} \sim \calD_{tr}} \left[  \log\left(1 +  \frac{\pi_\theta(y_l | x)}{\pi_\theta(y_w | x)} \right)  \right],
\end{eqnarray}
where $C$ is a constant independent of $\theta$.  Additionally, the third-to-last equality stems from the definition of how tuples $\{y_w,y_l,x\}$ are sampled.  In particular, for a given pair $\{y_1,y_2\}$, by definition a proportion $p^*(z=1| y_1, y_2,x)$ of the time $y_w = y_1$, while a proportion $p^*(z=0| y_1, y_2,x) = p^*(z=1|y_2,y_1,x)$ of the time $y_w = y_2$.  Hence
\begin{eqnarray}
 &&   p^*(z=1| y_1, y_2,x)\log p_\theta(z=1| y_1, y_2,x) ~+~p^*(z=1| y_2, y_1,x)\log p_\theta(z=1| y_2, y_1,x) \nonumber \\
 && ~~ \equiv ~~ \log p_\theta(z=1| y_w, y_l,x)
\end{eqnarray}
when the latter is averaged over the preference distribution.  \myendofproof

\section{Limitations} \label{app:limitations}
 As more of an analysis-driven contribution, our experiments on real-world data are limited to Figure \ref{fig:hh_results}.  Moreover, there are promising possibilities raised by pairing our contribution with prior work in new ways that we have not yet been explored.  One example is the potential use of REINFORCE in conjunction with modifications to the proposed $\ell_{\tiny \mbox{TYPO}}$ loss.

 \section{Broader Impacts} \label{app:impact}

Aligning the output of LLMs with human preferences has obvious, well-documented benefits.  However, there nonetheless remains the risk that tools designed to improve LLM responses could be repurposed for nefarious aims.  For example, preference data labels could potentially be modified to train models, using preference losses such as ours, that intentionally produce toxic content.
 
 % , such as 
 
 % Our work is framed as more of an analysis driven contribution.  In this vein, there are promising empirical directions to explore, such as the use of REINFORCE in Section \ref{sec:new_objectives} that we defer to future work.  
 
 %We also discuss a wide variety of interesting recent papers in Section \ref{sec:related_work} that are not directly addressed w.r.t.~our analysis. 

% \clearpage
% \input{checklist}

\end{document}